\documentclass[a4paper,fleqn]{cas-sc}

\usepackage[authoryear]{natbib}

\def\tsc#1{\csdef{#1}{\textsc{\lowercase{#1}}\xspace}}
\tsc{WGM}
\tsc{QE}
\tsc{EP}
\tsc{PMS}
\tsc{BEC}
\tsc{DE}

\usepackage{epigraph}
\usepackage{longtable}
\usepackage{setspace}

\usepackage{textcomp}

\usepackage{multirow}

\usepackage{color}

\definecolor{WordGreen}{RGB}{100, 136, 40}
\definecolor{WordDarkGrey}{RGB}{82, 82, 82}
\definecolor{WordRed}{RGB}{192, 80, 77}
\definecolor{WordBlue}{RGB}{0, 122, 192}

\definecolor{WordLightBlue}{RGB}{218, 238, 243}
\definecolor{WordLightGreen}{RGB}{234, 241, 221}

\definecolor{WordFillGreen}{RGB}{194, 214, 155}
\definecolor{WordFillRed}{RGB}{252, 214, 182}
\definecolor{WordFillGray}{RGB}{217, 217, 217}

\UseRawInputEncoding
\usepackage{times}
\usepackage{epsfig}
\usepackage{amssymb}
\usepackage[ruled,vlined,linesnumbered]{algorithm2e}

\usepackage{comment}
\usepackage{array, makecell} 
\usepackage{lettrine}
\usepackage{url} 
\usepackage{lscape}
\usepackage{cuted}
\usepackage{tabularx}
\usepackage{colortbl}
\usepackage{hhline}
\usepackage{vcell}
\usepackage{longtable}
\usepackage{vcell}
\usepackage{rotating}
\usepackage{pdflscape}
\usepackage{sidecap}
\usepackage{neuralnetwork}
\usepackage[export]{adjustbox} 
\usepackage[accsupp]{axessibility} 
\usepackage{amsfonts,bm}
\usepackage{physics}
\usepackage{wrapfig}
\usepackage{caption}
\usepackage{color,soul}
\usepackage[frozencache,cachedir=minted-cache]{minted}
\usepackage{mathtools}
\usepackage{amsthm}

\usepackage{acronym}
\acrodef{FCN}[FCN]{Fully Convolutional Network}
\acrodef{GAME}[GAME]{Grid Average Mean Absolute Error}
\acrodef{DL}[DL]{Deep Learning}
\acrodef{DNN}[DNN]{Deep Neural Network}
\acrodef{ML}[ML]{Machine Learning}
\acrodef{CV}[CV]{Computer Vision}
\acrodef{AI}[AI]{Artificial Intelligence}
\acrodef{CNN}[CNN]{Convolutional Neural Network}
\acrodef{RNN}[RNN]{Recurrent Neural Network}
\acrodef{GAN}[GAN]{Generative Adversarial Network}
\acrodef{JCU}[JCU]{James Cook University}
\acrodef{MAE}[MAE]{Mean Average Error}
\acrodef{MAP}[mAP]{Mean Average Precision}
\acrodef{CA}[CA]{Classification Accuracy}
\acrodef{LCFCN}[LCFCN]{Localization-based Counting loss Fully Convolutional Network}
\acrodef{IoT}[IoT]{Internet of Things}
\acrodef{MLP}[MLP]{Multi-Layer Perceptrons}
\def\q{{\mathbf q}}				
\def\rvy{{\mathbf{y}}}
\def\ervy{{\textnormal{y}}}

\usepackage{xspace}

\newcommand{\ce}{O}

\usepackage[capitalize]{cleveref}
\crefname{section}{Sec.}{Secs.}
\Crefname{section}{Section}{Sections}
\Crefname{table}{Table}{Tables}
\crefname{table}{Tab.}{Tabs.}
\usepackage{booktabs}
\usepackage{arydshln}
\usepackage{listings}
\definecolor{codegreen}{rgb}{0,0.6,0}
\definecolor{codegray}{rgb}{0.5,0.5,0.5}
\definecolor{codepurple}{rgb}{0.58,0,0.82}
\definecolor{backcolour}{rgb}{0.95,0.95,0.92}
\definecolor{cleacolorr}{rgb}{1,1,1}

\lstdefinestyle{mystyle}{
    backgroundcolor=\color{cleacolorr},   
    commentstyle=\color{codegreen},
    keywordstyle=\color{magenta},
    numberstyle=\tiny\color{codegray},
    stringstyle=\color{codepurple},
    basicstyle=\ttfamily\footnotesize,
    breakatwhitespace=false,         
    breaklines=true,                 
    captionpos=b,                    
    keepspaces=true,                 
    numbers=left,                    
    numbersep=5pt,                  
    showspaces=false,                
    showstringspaces=false,
    showtabs=false,                  
    tabsize=2
}

\usepackage[most]{tcolorbox}
\newtcolorbox[auto counter]{pabox}[2][]{%
colback=blue!5!white,colframe=blue!75!black,fonttitle=\bfseries,
title=Box~\thetcbcounter: #2,#1}

\usepackage{tikz}

\usepackage{graphicx}
\graphicspath{{./Graphics/}}
\DeclareGraphicsExtensions{.pdf,.jpeg,.png,.jpg}

\usepackage{amsmath}

\usepackage{array}

\usepackage{url}

\begin{document}

\let\WriteBookmarks\relax
\def\floatpagepagefraction{1}
\def\textpagefraction{.001}

\shorttitle{Weed Contrastive Learning through Visual Representation}

\shortauthors{Saleh et~al.}

\title [mode = title]{WeedCLR: Weed Contrastive Learning through Visual Representations with Class-Optimized Loss in Long-Tailed Datasets}                      

%
\author[1]{Alzayat Saleh}[orcid=0000-0001-6973-019X]
\fnmark[1]
\credit{Conceptualisation, Data Curation, Data Analysis, Software Development, DL Algorithm Design, Visualization, Writing original draft. }


\author[2]{Alex Olsen}
\credit{Data Curation, Reviewing/editing the draft.}

\author[2]{Jake Wood}
\credit{Reviewing/editing the draft. }

\author[1]{Bronson Philippa}
\credit{Reviewing/editing the draft.}

\author[1]{Mostafa~Rahimi~Azghadi}[                        orcid=0000-0001-7975-3985]
\cormark[1]
\ead{mostafa.rahimiazghadi@jcu.edu.au}
\credit{Conceptualisation, Data Curation, Data Analysis, Reviewing/editing the draft.}

\affiliation[1]{organization={College of Science and Engineering, James Cook University},
    city={Townsville},
    postcode={4814}, 
    state={QLD},
    country={Australia}}


\affiliation[2]{organization={AutoWeed Pty Ltd},
    city={Townsville},
    postcode={4814}, 
    state={QLD},
    country={Australia}}

\cortext[cor1]{Corresponding author}

\fntext[fn1]{This is the first author footnote.}

\begin{abstract}
Image classification is a crucial task in modern weed management and crop intervention technologies. However, the limited size, diversity, and balance of existing weed datasets hinder the development of deep learning models for generalizable weed identification. In addition, the expensive labelling requirements of mainstream fully-supervised weed classifiers make them cost- and time-prohibitive to deploy widely, for new weed species, and in site-specific weed management. 
This paper proposes a novel method for Weed Contrastive Learning through visual Representations (WeedCLR), that uses class-optimized loss with Von Neumann Entropy of deep representation for weed classification in long-tailed datasets. WeedCLR leverages self-supervised learning to learn rich and robust visual features without any labels and applies a class-optimized loss function to address the class imbalance problem in long-tailed datasets. WeedCLR is evaluated on two public weed datasets: CottonWeedID15, containing 15 weed species, and DeepWeeds, containing 8 weed species. WeedCLR achieves an average accuracy improvement of $ 4.3\%$ on CottonWeedID15 and $5.6\%$ on DeepWeeds over previous methods. It also demonstrates better generalization ability and robustness to different environmental conditions than existing methods without the need for expensive and time-consuming human annotations. These significant improvements make WeedCLR an effective tool for weed classification in long-tailed datasets and allows for more rapid and widespread deployment of site-specific weed management and crop intervention technologies.
\end{abstract}


\maketitle
 

\section{Introduction}\label{secintro}

Weed management is a critical issue in agriculture, as weeds compete with crops for resources and can significantly reduce crop yields \citep{Rai2023ApplicationsReview, Xu2023PrecisionChallenges}. Existing weed management relies heavily on broadcast application of herbicides to entire paddocks to control weeds, including in areas of paddocks which do not contain weeds. This over-application of herbicide has both negative economic and environmental effects that could be minimised with site-specific weed control. In the past two decades, new approaches for site-specific weed management have been developed to detect weeds and only apply herbicide where it is needed. WeedIT \citep{weed-it}  and WeedSeeker \citep{trimble}  uses near-infrared (NIR) sensing technology to detect weeds which are limited to weed control in-fallow application, or \textit{green-on-brown} spraying. While deep learning based approaches have shown promise for in-crop application \citep{Arsa2023Eco-friendlyNetworks, Dang2023YOLOWeeds:Systems}, or \textit{green-on-green} spraying. The latter approaches are on the cusp of achieving significant uptake in the commercial arena \citep{Coleman2022WeedSystems}. However, supervised learning is the predominant deep learning method for these approaches, which requires human involvement to annotate the presence of weeds within each image. This annotation process, being highly time-consuming, stands as a formidable barrier to the widespread adoption of these approaches. Furthermore, the specialized knowledge required for accurately identifying weed species in images, particularly within specific cropping systems, makes it impractical to outsource this work to online paid annotation platforms. This is underscored by the fact that even among trained plant consultants, a noteworthy 12\% error rate has been reported \citep{dyrmann2016evaluation}. Self-supervised learning techniques have the ability to group clusters within datasets without the need for human annotation. There is limited research on their use for weed recognition \citep{Coleman2022WeedSystems}. Another challenge for deep learning approaches is that the ideal use case for site specific weed management is when weed pressure is low, which maximises the reduction in herbicide usage. Consequently, a small proportion of target weed species creates a small sample of target images in the dataset, otherwise known as long-tailed datasets. This class imbalance is an obstacle that must be overcome. Below, we explain how we extend the state-of-the-art to develop WeedCLR to address the two aforementioned challenges. 

Self-supervised learning methods have shown great potential to align the embedding vectors of augmented views of a training instance. One such technique is contrastive learning \citep{Chen2020b}, which compares training samples by treating each sample as its own class. This is typically achieved through the use of the InfoNCE contrastive loss \citep{vandenOordDeepMind2018RepresentationCoding}, which brings representations of positive pairs of examples closer together in the embedding space while pushing representations of negative pairs further apart. However, this approach is known to require a large number of negative samples.
In contrast, non-contrastive methods do not rely on explicit negative samples and include techniques such as clustering-based approaches \citep{Caron2020UnsupervisedAssignments}, redundancy reduction methods \citep{Zbontar2021BarlowReduction}, and methods that utilize specialized architecture design \citep{grill2020bootstrap, Xie2022SimMIM:Modeling}. These methods have shown promising results in various applications and continue to be an active area of research.

Many self-supervised methods manipulate the input data to extract a supervised signal in the form of a pre-designed (pretext) task. One such task of particular interest is the jigsaw puzzle task, which has been encoded as an invariant for contrastive learning. In contrast to this approach, we have used a \texttt{multi-crop} strategy that simply samples multiple random crops of the input image, with two different sizes: a standard size and a smaller one. This enables simultaneous learning of representations and cluster assignments in an end-to-end fashion, providing a more efficient and effective method for self-supervised learning.

Theoretical analysis of self-supervised learning has attempted to understand the underlying dynamics of these methods, which have shown success in learning useful representations and outperforming their supervised counterparts in several downstream transfer learning benchmarks \citep{Chen2020b}. Despite their success, the mechanics of these methods remain somewhat obscure and poorly understood. Several studies have provided theoretical evidence that representations learned via contrastive learning are useful for downstream tasks \citep{Arora2019ALearning, Lee2020PredictingLearning, Tosh2021ContrastiveModels}. Additionally, it has been explained by \citep{Tian2021UnderstandingPairs} why non-contrastive learning methods such as BYOL \citep{grill2020bootstrap} and SimSiam \citep{Xie2022SimMIM:Modeling} work, with the alignment of eigenspaces between the predictor and its input correlation matrix playing a key role in preventing complete collapse.

In terms of implicit regularization, it has been theoretically demonstrated that gradient descent will drive adjacent matrices to align in a linear neural network setting \citep{Ji2019GradientNetworks}. Under the assumption of aligned matrices \citep{Gunasekar2018ImplicitFactorization}, it has been shown that gradient descent can derive nuclear-norm minimization solutions. This concept has been extended to deep linear networks by \citep{Arora2019ImplicitFactorization}, with both theoretical and empirical evidence demonstrating that deep linear networks can derive low-rank solutions. In general, over-parametrized neural networks tend to find flatter local minima \citep{Saxe2019ANetworks, Neyshabur2019TowardsNetworks, Barrett2021ImplicitRegularization}.
In this work, we use a regularization method, Von Neumann Entropy (VNE) of deep representation. This regularizer can effectively control not only VNE but also other theoretically related properties, including decorrelation and rank. Our method considers the eigenvalue distribution of the autocorrelation matrix, which describes the correlation between different features in the representation.

The main objective of this study is to develop an efficient and effective weed classification method by employing our above-mentioned deep learning contributions in self-supervised learning and regularization. To achieve this objective, we developed WeedCLR, for weed classification in long-tailed datasets using a self-visual features learning approach. WeedCLR uses a class-optimized loss function to improve classification accuracy and leverages the power of self-supervised learning to extract meaningful visual features from images of weeds without the need for any human annotation  \citep{Magistri2023FromRobotics, Espejo-Garcia2023Top-tuningIdentification}. These features are then used to train a classifier that can accurately distinguish between different weed species. We evaluate our method on two public weed datasets: CottonWeedID15 \citep{Chen2022cotton} and DeepWeeds \citep{Olsen2019DeepWeed}, and demonstrate its effectiveness in improving classification accuracy compared to state-of-the-art self-supervised learning approaches.

Our approach includes the use of Von Neumann Entropy (VNE) to optimize the representation space, which helps us to achieve a more desirable representation that avoids dimensional collapse and produces more useful embeddings. We also introduce a regularization that encourages the model to assign labels uniformly across all classes, preventing degenerate solutions where all labels are assigned to a single class. Additionally, our method aims to optimize the model behavior for classes other than the correct class by maximizing the likelihood of the correct class while neutralizing the probabilities of the incorrect classes. This helps us to achieve better classification accuracy and improve the performance of our model in practical long-tailed weed datasets.

Specifically, our method is a simple yet effective self-supervised single-stage end-to-end classification and representation learning technique for weed classification. It does not require any form of pre-training, expectation-maximization algorithm, pseudo-labeling, or external clustering, unlike previous unsupervised classification works \citep{Chen2020b, Caron2020UnsupervisedAssignments}. Additionally, our approach does not require a memory bank, a second network (momentum), external clustering, stop-gradient operation, or negative pairs, unlike previous unsupervised representation learning works \citep{Chen2020ImprovedLearning, grill2020bootstrap, Caron2021EmergingTransformers}.

Our main contributions are summarized as follows:
\begin{enumerate}
    \item We present, to the best of our knowledge, the first self-supervised weed classification technique for long-tailed weed datasets. This has significant benefits for practical weed classification and management technologies. 
    \item Our novel method, Weed Contrastive Learning through visual Representations (WeedCLR) leverages the power of multi-crop strategy with self-supervised learning to extract meaningful visual features from images of weeds, without the need for human annotation. The features are then used to train a classifier that can accurately distinguish between different weed species.
    \item 
    WeedCLR utilizes a class-optimized loss function to improve classification accuracy in long-tailed datasets, which are very common in real-world weed datasets.
    \item WeedCLR includes the use of Von Neumann Entropy (VNE) to optimize the representation space, which helps it achieve a more desirable representation that avoids dimensional collapse and produces more useful embeddings.
    \item WeedCLR introduces a regularization that encourages the model to assign labels uniformly across all classes, preventing degenerate solutions where all labels are assigned to a single class.
    \item WeedCLR optimizes the model behavior for classes other than the correct class by maximizing the likelihood of the correct class while neutralizing the probabilities of the incorrect classes. This helps it achieve better classification accuracy and performance.    
    \item We evaluate WeedCLR on two public weed datasets: CottonWeedID15 \citep{Chen2022cotton} and DeepWeeds \citep{Olsen2019DeepWeed}, and demonstrate its effectiveness in improving classification accuracy compared to state-of-the-art self-supervised learning approaches.
\end{enumerate}

The rest of this paper is organized as follows: 
Section \ref{secmethod} presents our proposed WeedCLR approach in detail. Section \ref{secExperiments}, presents the experimental setup and implementation details of WeedCLR. Section \ref{ssecrslt} evaluates the performance of our approach on two benchmark datasets and compares it to several state-of-the-art methods. Section \ref{secabl} investigates the effect of different factors on the DeepWeeds and CottonWeedID datasets. Section \ref{secdisc}  discusses the limitations of our approach and future work. Finally, Section \ref{secconc} summarizes the main findings of this paper and outlines the potential impact of our work.

\section{Method} \label{secmethod}
Despite the recent successes of self-supervised learning (SSL) methods, they are not as effective when dealing with unbalanced datasets. This is because many SSL methods rely on a hidden uniform prior, which distributes the data uniformly in the representation space. This causes the model to learn the most discriminative features in a given mini-batch. When the data is evenly distributed across classes, the most discriminative features that the model will learn will be class-specific. However, when using imbalanced data, the most discriminative features inside the mini-batch might not be the class anymore but more low-level information, thus decreasing performance on downstream classification tasks.

The most common practice for pretraining SSL models is to use curated datasets such as ImageNet and PASS, which are usually class-balanced and contain images with a single object prominently featured in the centre. However, these datasets are not always representative of the data found in the wild, such as weed datasets, which are often unbalanced.

In this section, we present our WeedCLR method for training a classifier that can accurately classify two different augmented views of the same image sample. Our goal is to train a classifier that can classify the two views similarly while avoiding degenerate solutions. By doing so, we aim to improve the performance of SSL methods on unbalanced weed datasets. However, our proposed technique can be applied to any other image datasets, especially unbalanced ones.

\begin{figure*}[!t]
\centering
\includegraphics[width=0.98\textwidth]{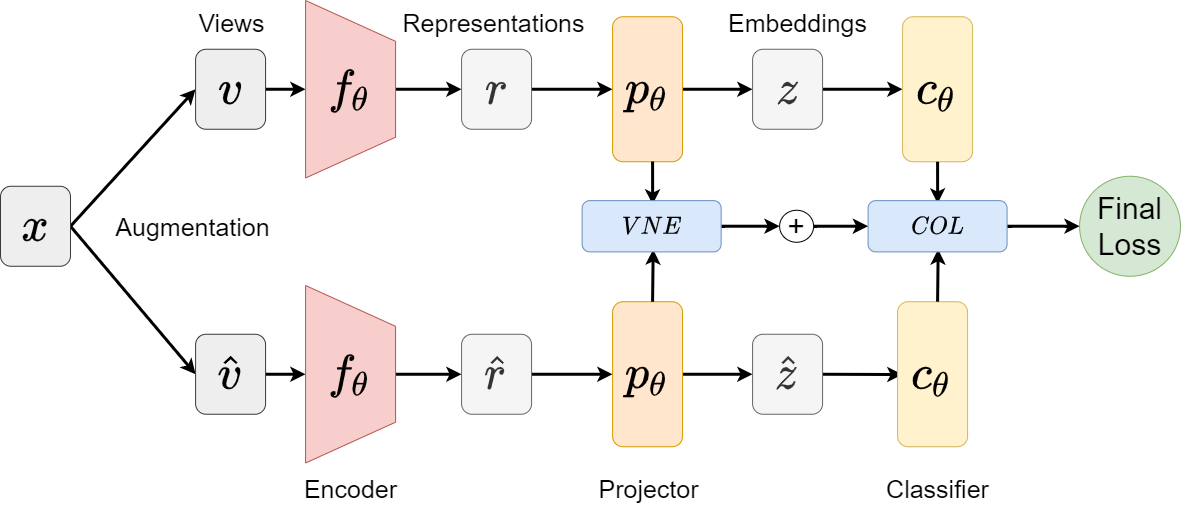}
\caption{Illustration of the architecture of our proposed WeedCLR model, which is designed for visual representation learning. The WeedCLR architecture processes two augmented views of the same image through a shared network comprised of an Encoder $f_\theta$ (e.g. CNN) and a Classifier (e.g. Projection MLP + linear classification head). The eigenvalue distribution of the two views (Representations) is optimized using Von Neumann Entropy (VNE) to avoid dimensional collapse, and the Class-Optimized Loss (COL) is minimized to promote the same class prediction while avoiding degenerate solutions by asserting a uniform prior on class predictions. The resulting model learns representations and discovers the underlying classes in a single-stage end-to-end unsupervised manner, allowing for efficient and effective classification of images in an unsupervised manner.}
\label{fig:3}
\end{figure*}

A naive approach to this problem would be to minimize a cross-entropy loss function, as shown in Equation \ref{eq:ell_naive}. 
\begin{equation}
    \tilde{\ell}(x_1, x_2) = -\sum_{y \in [C]}{p(y|x_2)\log p(y|x_1)},
    \label{eq:ell_naive}
\end{equation}
where $p(y|x)$ represents the probability of class $y$ given input $x$, calculated as a row softmax of the matrix of logits $\mathcal{S}$ in one-hot encoded representation. This matrix is produced by our model (encoder + classifier), see \cref{fig:3}, for all classes (represented by columns) and batch samples (represented by rows). 

However, without additional regularization, this approach quickly converges to a degenerate solution, (\textit{i.e.}, the network predicts a constant $y$ regardless of the input $x$). To address this issue, we propose optimizing our model's representation space by incorporating Von Neumann Entropy (VNE), and introducing a novel loss function called Class-optimized Loss (COL), as shown in \cref{fig:3}. In the following subsections, we will explain the Representation Space Optimization and our proposed Class-optimized Loss (COL) in more detail.

\begin{figure*}[t]
\centering
\includegraphics[width=0.99\textwidth]{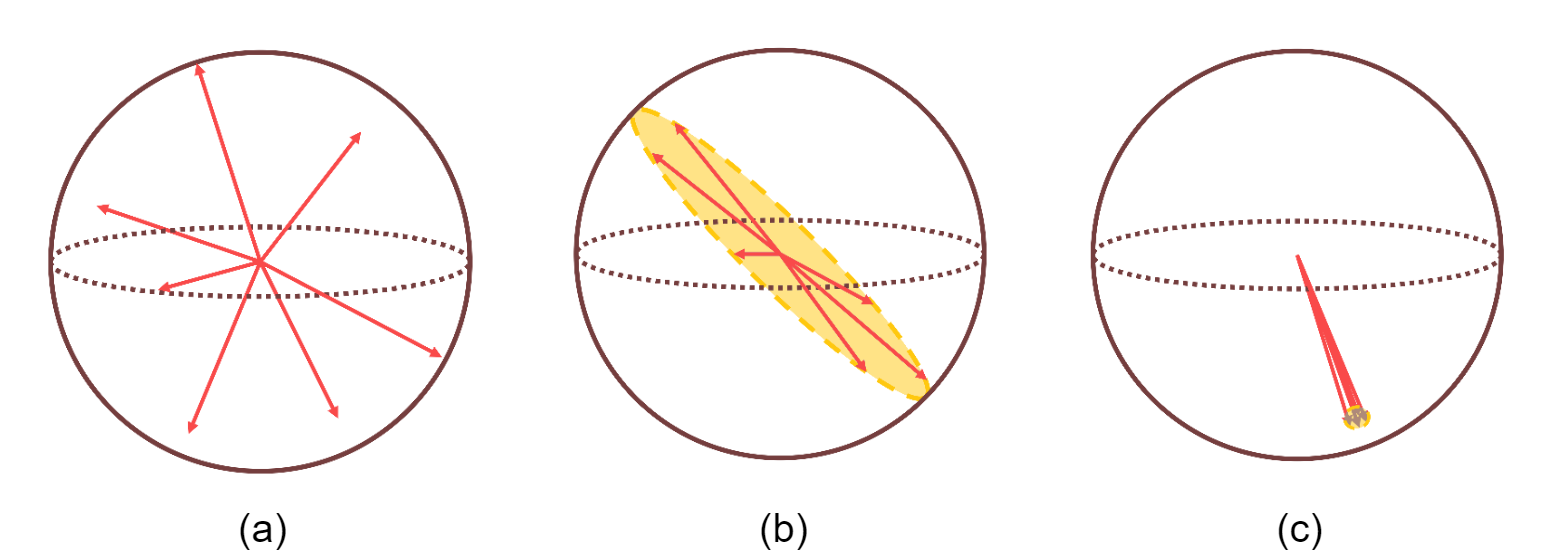}
\caption{Illustratation of the representation collapse problem. (a) Ideal representation space (b) Dimensional collapse (c) Complete collapse. For dimensional collapse, the embedding vectors only span a lower dimensional space. For complete collapse, the embedding vectors collapse to the same point.}
\label{fig:4}
\end{figure*}

\subsection{Representation Space Optimization}

Self-supervised learning methods  learn useful representations by minimizing the distances between embedding vectors from augmented images (\cref{fig:3}). Without additional regularization, this would result in a collapsed solution where the produced representation becomes constant (\cref{fig:4}c). However, some contrastive methods prevent complete collapse via the negative term that pushes embedding vectors of different input images away from each other \citep{Chen2020ImprovedLearning, grill2020bootstrap, Caron2021EmergingTransformers}. While contrastive methods prevent complete collapse, they still experience a dimensional collapse in which the embedding vectors occupy a lower-dimensional subspace than their dimension (\cref{fig:4}b).

To achieve the ideal representation space (\cref{fig:4}a), we propose a method that uses Von Neumann Entropy (VNE), as shown in  \cref{fig:3}, to control the eigenvalue distribution of the autocorrelation matrix \citep{kim2023vne}. 
By optimizing the representation space using VNE, we can achieve a more desirable representation that avoids dimensional collapse and produces more useful embeddings.

Von Neumann Entropy (VNE) is a mathematical formulation used to manipulate representation properties and is a measure of the diversity or spread of the eigenvalues of the autocorrelation matrix, which is an important property of a representation. By regularizing the VNE of the representation, the eigenvalue distribution can be effectively manipulated to improve the quality of the representation.

The autocorrelation matrix of the representation, defined as $\mathcal{Z}_{\text{auto}}$,  is a mathematical construct that describes the correlation between different components of a representation.
The properties of the autocorrelation matrix are closely related to various representation properties, such as decorrelation, and rank. Decorrelation refers to the process of removing the correlation between different components of a representation, which can improve the quality of the representation.  Rank is a measure of the dimensionality of the representation, which can affect its expressiveness and generalization ability.
For a given mini-batch of $N$ samples, the representation matrix can be denoted as $\bm{H} = [\bm{h}_1, \bm{h}_2, ..., \bm{h}_N]^T \in \mathbb{R}^{N \times d}$, where $d$ is the size of the representation vector. We assume $L_2$-normalized representation vectors satisfying $||\bm{h}_i||_2=1$, as in previous studies \citep{kim2023vne}. 
This assumption is important because it ensures that the representation vectors have a consistent scale and are not affected by differences in magnitude.
By normalizing the vectors in this way, the method can focus on the distribution of the vectors rather than their absolute values.
The autocorrelation matrix of the representation is then defined as: 

\begin{equation}
\label{equ:cauto}
\mathcal{Z}_{\text{auto}} \triangleq \sum_{i=1}^N\frac{1}{N}\bm{h}_i\bm{h}_i^T=\frac{1}{N}\bm{H}^T\bm{H},
\end{equation}
where $\boldsymbol{h}_i$ is a set of vectors,  $i$ ranges from $1$ to $N$. 

The autocorrelation matrix in \cref{equ:cauto}  is denoted by $\mathcal{Z}_{\text{auto}}$, and is calculated by taking the outer product of each vector with itself, and then averaging over all $N$ vectors. The outer product of a vector $\boldsymbol{h}_i$ with itself is given by $\boldsymbol{h}_i \boldsymbol{h}_i^T$, where $\boldsymbol{h}_i^T$ denotes the transpose of $\boldsymbol{h}_i$. The sum of all such outer products is then divided by $N$ to obtain the average.

Alternatively, the expression for $\mathcal{Z}_{\text{auto}}$ can be written in terms of the matrix $\boldsymbol{H}$, where each column of $\boldsymbol{H}$ corresponds to a vector $\boldsymbol{h}_i$. Specifically, $\mathcal{Z}_{\text{auto}}$ can be written as $\boldsymbol{H}^T \boldsymbol{H} / N$, where $\boldsymbol{H}^T$ denotes the transpose of $\boldsymbol{H}$. This expression is useful for computing the autocorrelation matrix efficiently using matrix multiplication.

In the extreme case where $\mathcal{Z}_{\text{auto}} \rightarrow c \cdot I_d$, where $c$ is an adequate positive constant, $I_d$ a diagonal matrix, the eigenvalue distribution of $\mathcal{Z}_{\text{auto}}$ becomes perfectly uniform.
This means that each feature in the representation contributes equally to the overall variance, which can be beneficial for self-supervised contrastive learning methods.
The constant $c$ is chosen to be an adequate positive value, which ensures that the matrix remains positive definite and invertible.
This results in a  full-rank  representation and can prevent   dimensional collapse in contrastive learning.
Regularizing $\mathcal{Z}_{\text{auto}}$ is of great interest because it permits a simple implementation as a penalty loss. 

VNE of autocorrelation is defined as the Shannon entropy over the eigenvalues of $\mathcal{Z}_{\text{auto}}$, as shown in the equation below:

\begin{equation}
\label{equ:vne}
S(\mathcal{Z}_{\text{auto}}) \triangleq -\sum_j \lambda_{j} \log{\lambda_{j}}.
\end{equation}
where, $S$ denotes the Shannon entropy, which is a measure of the amount of uncertainty or randomness in a representation space.
The subscript $j$ represents the $j$-th eigenvalue of the autocorrelation matrix, denoted by $\lambda_j$.
The equation calculates the sum of the product of each eigenvalue and its natural logarithm, which is then multiplied by $-1$ to obtain the Von Neumann Entropy.

Implementing VNE regularization is simple. When training our model, we  subtract $\alpha \cdot S(\mathcal{Z}_{\text{auto}})$ from the main loss $\mathcal{L}$.

$$\mathcal{L}_{\text{opt}}=\mathcal{L}  - \alpha \cdot S(\mathcal{Z}_{\text{auto}}),$$ 
where $\mathcal{L}_{\text{opt}}$ is the loss function that is being optimized,  $\mathcal{L}$ is the original loss function without any regularization, $\alpha$ is a hyperparameter that controls the strength of the regularization,  $S\left(\mathcal{Z}_{\text {auto }}\right)$ is the Von Neumann Entropy of the autocorrelation matrix $\mathcal{Z}_{\text {auto}}$ of the learned representation.
The regularization term $-\alpha \cdot S\left(\mathcal{Z}_{\text {auto}}\right)$ encourages the learned representation to have a more structured eigenvalue distribution, which can improve its quality.

In the following subsection, we will explain  our proposed Class-optimized Loss (COL) function.

\subsection{Class-optimized Loss (COL)}

Our Class-Optimized Loss (COL) is composed of two components: The Uniform Prior Loss for the correct class and the Optimized Loss for the incorrect classes.

\subsubsection{The Uniform Prior Loss}
The first component of our Class-Optimized Loss (COL) is the modified cross-entropy loss for the correct class and is derived by applying Bayes theorem and the law of total probability, resulting in equations \ref{eq:full_bayes} and \ref{eq:total_probability}, which are used to compute the probability of a label given an augmented sample.

\begin{equation}
    p(y|v_2) = \frac{p(y)p(v_2|y)}{p(v_2)} = \frac{p(y)p(v_2|y)}{\sum_{\tilde{y} \in [C]}{p(v_2|\tilde{y}) p(\tilde{y})}},
    \label{eq:full_bayes}
\end{equation}
\begin{equation}
    p(y|v_1) = \frac{p(y)p(y|v_1)}{p(y)} = \frac{p(y)p(y|v_1)}{\sum_{\tilde{v_1} \in B_1}{p(y|\tilde{v}_1) p(\tilde{v}_1)}},
    \label{eq:total_probability}
\end{equation}
where  $C$ is the number of classes, such that two augmented views ($v_1, v_2$) of the same sample are classified similarly, $B$ represents a batch of $N$ samples, with $B_1$ denoting the first set of augmentations for the samples in $B$. The term $p(v|y)$ refers to a column softmax of the matrix of logits $\mathcal{S}$ mentioned earlier.

In   equation \ref{eq:full_bayes}, the probability of $y$ given $v_2$  is computed using Bayes theorem. It is equal to the product of the prior probability of $y$ and the likelihood of $v_2$ given $y$, divided by the evidence or marginal probability of $v_2$. The evidence is calculated by summing the product of the likelihood of $v_2$ given each possible label and the prior probability of each label over all possible labels.
In   equation \ref{eq:total_probability}, the probability of $y$ given $v_1$ is also computed using the Bayes theorem. It is equal to the product of the prior probability of $y$ and the likelihood of $y$ given $v_1$, divided by the prior probability of $y$. The prior probability of $y$ cancels out in the numerator and denominator, so it is not necessary to compute it explicitly. The likelihood of $y$ given $v_1$ is calculated by summing the product of the probability of $y$ given each possible value of $v_1$ and the probability of each value of $v_1$ over all possible values of $v_1$.

Assuming a uniform prior for $p(y)$ and a uniform distribution for $p(v_1)$, the proposed loss function is  mathematically equivalent to the naive cross-entropy loss under the assumption of uniform $p(y)$ and $p(v)$.  
Our proposed loss function, shown in   \cref{eq:ce_ours}, takes into account the prior probabilities of the classes and samples. 
\begin{equation}
    \small
    \ell(v_1, v_2) = -\sum_{y \in [C]}{\frac{p(v_2|y)}{\sum_{\tilde{y}}{p(v_2|\tilde{y})}}\log \bigg(\frac{N}{C}\frac{p(y|v_1)}{\sum_{\tilde{v}_1}{p(y|\tilde{v}_1)}}\bigg)},
    \label{eq:ce_ours}
\end{equation}
where $p(y)$ and $p(\tilde{y})$ cancel out in \cref{eq:full_bayes}, and $p(y)/p(\tilde{v}_1)$ becomes $\frac{N}{C}$ in \cref{eq:total_probability}. 

This cancellation occurs in   \cref{eq:full_bayes} because both terms represent the probability of the same label for two different augmented views of the same sample.
In   \cref{eq:total_probability}, the ratio of $p(y)$ and $p(\tilde{v}_1)$ is simplified to $\frac{N}{C}$, where $N$ is the total number of samples and $C$ is the number of classes. This simplification is possible because $p(\tilde{v}_1)$ represents the probability of the augmented view of the sample, which is assumed to be uniformly distributed across all samples. Therefore, $p(\tilde{v}_1)$ can be approximated as $\frac{1}{N}$, and the ratio $\frac{p(y)}{p(\tilde{v}_1)}$ becomes $Np(y)$.
Since the loss function is defined as the negative log-likelihood of the predicted labels, this simplification results in a term of $-\log\left(\frac{N}{C}\right)$ in the loss function. This term acts as a regularization that encourages the model to assign labels uniformly across all classes, preventing degenerate solutions where all labels are assigned to a single class. This means that a solution with an equal distribution of data among all classes by  asserting a uniform prior on the standard cross-entropy loss function is considered an optimal solution.

\begin{figure*}[!t]
\centering
\includegraphics[width=0.98\textwidth]{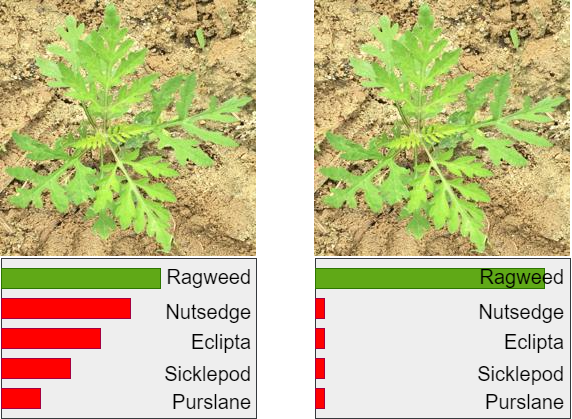}
\caption{Illustration of the top-5 predicted probabilities $\hat{\ervy}$ from two training paradigms for a sample image from the CottonWeedID15 \citep{Chen2022cotton}  dataset, with the ground-truth class being "Ragweed". The model used is ResNet-50. The left graph shows the predicted probabilities from the model trained with The Uniform Prior Loss only, while the right graph shows the predicted probabilities from the model trained with our proposed Class-optimized Loss (COL), which includes both The Uniform Prior Loss and The Optimized Loss. Compared to the right graph, the model in the left graph is confused by other classes such as "Nutsedge" and "Eclipta", suggesting that it might be more susceptible to generalization issues and potentially adversarial attacks.}
\label{fig:9}
\end{figure*}

In practice, we employ a symmetric variant of this loss, as shown in Equation \ref{eq:L_sym}. Empirical evidence suggests that this variant yields better results.
\begin{equation}
    \mathcal{L} = \frac{1}{2}\bigg(\ell(v_1, v_2) + \ell(v_2, v_1)\bigg).
    \label{eq:L_sym}
\end{equation}

While our Uniform Prior Loss  (\cref{eq:ce_ours}) has shown improved results in avoiding degenerate solutions (see \cref{secabl}), it primarily exploits information from the correct class and largely ignores information from incorrect classes. This can be attributed to the fact that predicted probabilities other than $\tilde{y}$ are zeroed out during the dot product calculation with one-hot encoded $y$. As a result, model behavior for classes other than the correct class is not explicitly optimized (see \cref{fig:9}  left graph). Instead, their predicted probabilities are indirectly minimized when $\tilde{y}$ is maximized, given that probabilities must sum to $1$. This effect is more pronounced in datasets with imbalanced class distributions. To address this limitation, an optimized loss for incorrect classes has been proposed as an additional regularization to  our cross-entropy variant.

\subsubsection{The Optimized Loss}
The second component of our Class-Optimized Loss (COL) aims to explicitly optimize the model behavior for classes other than the correct class, by  maximising the likelihood of the correct class while neutralizing the probabilities of the incorrect classes.

 The Optimized Loss $\ce(\cdot)$  is defined as the average of sample-wise entropies over incorrect classes in a mini-batch, as shown in   \cref{eq:complement_cross_entropy}.

\begin{equation}
\label{eq:complement_cross_entropy}
\begin{aligned}
\ce(\hat{\rvy}_{\bar{c}}) 
&= \frac{1}{N}\sum_{i=1}^N \mathcal{H}(\hat{\rvy}_{i\bar{c}})\\
&= -\frac{1}{N}\sum_{i=1}^N \sum_{j=1, j\neq g}^{K} (\frac{\hat{\ervy}_{ij}}{1-\hat{\ervy}_{ig}})\log(\frac{\hat{\ervy}_{ij}}{1-\hat{\ervy}_{ig}})
\end{aligned}
\end{equation}
where  $\hat{\rvy}_{\bar{c}}$  is  the predicted probabilities of the  incorrect classes,  
$\mathcal{H}(\cdot)$ is the entropy function,
 $\rvy_i$ is one-hot vector representing the label of the $i$\textsuperscript{th} sample,   
 $\hat{\rvy}_i$ is the predicted probability for each class for the $i$\textsuperscript{th} sample,  
 $g$ is the index of the correct class,  
 $\ervy_{ij}$ or $\hat{\ervy}_{ij}$ is the $j$\textsuperscript{th} class (element) of $\rvy_i$ or $\hat{\rvy}_i$,  
$N$ and $K$  are  the total number of samples and the total number of classes.

\cref{eq:complement_cross_entropy} calculates the Optimized Loss, $\ce(\cdot)$, using a set of predicted probabilities, $\hat{\rvy}_{\bar{c}}$. 
The entropy function is denoted as $\mathcal{H}(\cdot)$, which is a measure of the uncertainty or randomness of a probability distribution. The sample-wise entropy is calculated by considering only the incorrect classes other than the correct class $g$. This means that the entropy is calculated based on the predicted probabilities of all classes except the correct class for a given sample. The predicted probability for each class, denoted by $\hat{\ervy}_{ij}$, is normalized by one minus the correct probability (i.e., $1-\hat{\ervy}_{ig}$). This normalization ensures that the predicted probabilities sum up to one.

The term $\frac{\hat{\ervy}_{ij}}{1-\hat{\ervy}_{ig}}$ can be interpreted as the predicted probability of observing class $j$ for the $i$-th sample, given that the correct class $g$ does not occur. Since entropy is maximized when events are equally likely to occur, optimizing on the incorrect entropy drives $\hat{\ervy}_{ij}$ to $\frac{(1-\hat{\ervy}_{ig})}{(K-1)}$, where $K$ is the total number of classes. This essentially neutralizes the predicted probability of incorrect classes as $K$ grows large. Maximizing the incorrect entropy ``flattens" the predicted probabilities of incorrect classes $\hat{y}_{j \ne g}$. This means that the predicted probabilities of incorrect classes are reduced, making it less likely for the neural net $h_\theta$ to make incorrect predictions (see \cref{fig:9} right graph).

We hypothesise that when the predicted probabilities of incorrect classes are neutralized, the neural net $h_\theta$ generalizes better. This is because it is less likely to have an incorrect class with a sufficiently high predicted probability to ``challenge" the correct class. In other words, by maximizing the incorrect entropy and ``flattening" the predicted probabilities of incorrect classes (see \cref{fig:9}), the neural net is able to make more accurate predictions and generalize better to new data.

\begin{figure*}[t]
\centering
\includegraphics[width=0.99\textwidth]{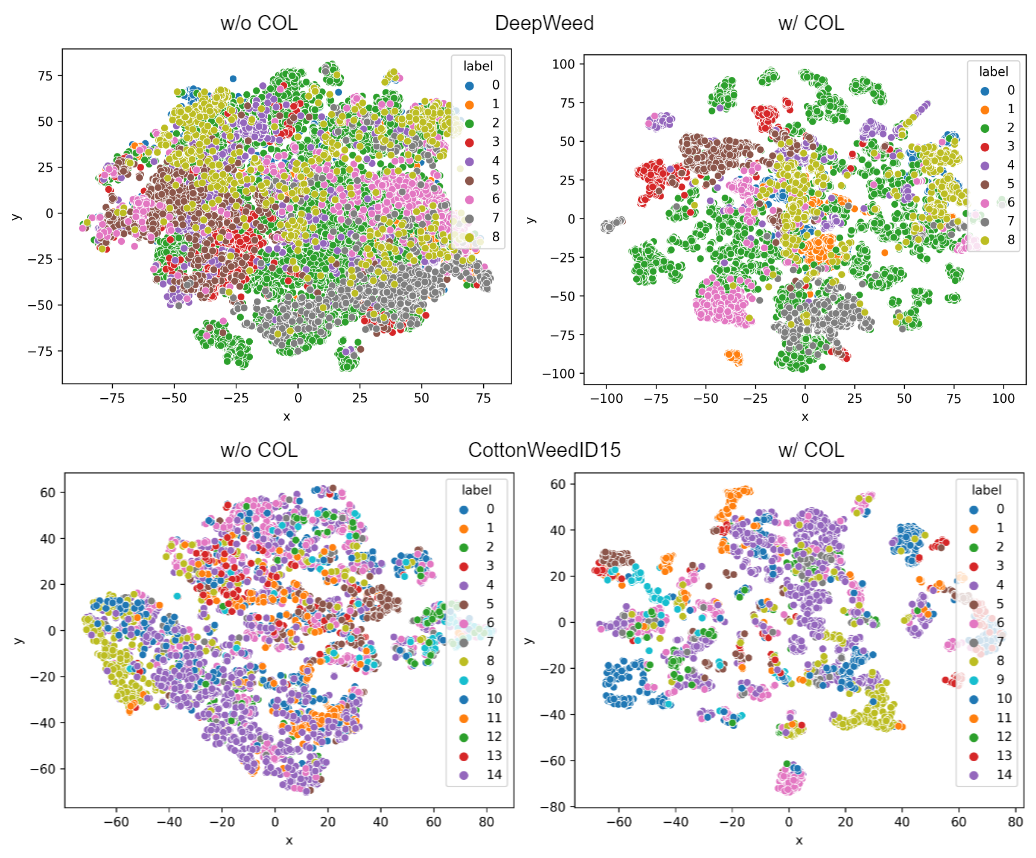}
\caption{
The embeddings for DeepWeeds and CottonWeedID15  test images obtained from two self-supervised training paradigms. On the left, the embeddings were obtained by training with cross-entropy only, while on the right, the embeddings were obtained by training with our COL. The model used is ResNet-50, and the embeddings represent the vector representation before feeding to the MLP and classification heads. The embeddings were projected to 2D vectors using t-SNE for visualization purposes. 
}
\label{fig:6}
\end{figure*}

\subsection{WeedCLR Model}
WeedCLR is a novel approach to self-supervised end-to-end classification learning that learns labels and representations simultaneously in a single-stage, end-to-end manner, as shown in \cref{fig:3}. This approach optimizes for same-class prediction of two augmented views of the same sample and employs a mathematically motivated variant of the cross-entropy loss with a uniform prior asserted on the predicted labels to prevent all labels from being assigned to the same class. WeedCLR explicitly optimizes model behavior for classes other than the correct class by maximizing the likelihood of the correct class while neutralizing the probabilities of the incorrect classes. WeedCLR is easy to implement and scalable. Unlike other popular unsupervised classification and contrastive representation learning approaches, it does not require pre-training, expectation-maximization, pseudo-labeling, external clustering, a second network, stop-gradient operation, or negative pairs.

\cref{alg:pseudo_code} describes a training procedure for our WeedCLR model, also illustrated in \cref{fig:3}. The algorithm takes as input a training dataset $\mathbf{D}$ and outputs the model parameters $\mathbf{\theta}$. The training process consists of several steps, which are repeated for a specified number of training steps $n_{train\_steps}$.

First, a mini-batch of data $\mathbf{X}$ is obtained from the training dataset $\mathbf{D}$ (line 3). This mini-batch is then augmented to produce two sets of data, $\mathbf{v}$ and $\hat{\mathbf{v}}$ (line 4). These augmented data are then passed through an MLP encoder to extract features $\mathbf{z}$ and $\hat{\mathbf{z}}$ (line 5). These features are optimized by $\mathcal{Z}_{\text{VNE}}$ which is calculated using equation \ref{equ:vne} (line 6). These features are then classified using a CLF function (classifier) to produce predicted class probabilities ${\rvy}_{\bar{c}}, \hat{\rvy}_{\bar{c}}$ (line 7).

Next, the COL loss term $\mathcal{L}_{\text{COL}}$ is  calculated using equation \ref{eq5} (line 8). The final loss is then calculated as the sum of the VNE and COL loss terms (line 9), and the model parameters are updated using this final loss and an optimizer (line 10).

\vspace{1.5cm}
\begin{algorithm} 
\SetKwInOut{Input}{input}
\SetKwInOut{Output}{output}
\Input{Training dataset, {$\mathbf{D}=\{\mathbf{X}_{i}, \cdots, \mathbf{X}_{n}\}^{N}_{i=1} $}}
\Output{Model parameters, \{$\mathbf{\theta}_{1}$, $\cdots$, $\mathbf{\theta}_{n_{layers}}$\}}
 initialization\;
  \For{$t \leftarrow 1$ $\emph{\textbf{to}}$ $n_{train\_steps}$}{
        $\mathbf{X}$ $\gets$ mini\_batch($\mathbf{D}, t$)\;
        $\mathbf{v}, \hat{\mathbf{v}}$ $\gets$ augmentation($\mathbf{X}$)\;
        $\mathbf{z}, \hat{\mathbf{z}}$ $\gets$ MLP(encoder($\mathbf{v}, \hat{\mathbf{v}}$))\;
        $\mathcal{Z}_{\text{VNE}} \gets  - \alpha \cdot S(\mathcal{Z}_{\text{auto}}).$  \Comment{  \cref{equ:vne}}\;
        ${\rvy}_{\bar{c}}, \hat{\rvy}_{\bar{c}}$ $\gets$ CLF($\mathbf{z}, \hat{\mathbf{z}}$)\;
        $\mathcal{L}_{\text{COL}}$ $\gets$ $\mathcal{H}({\rvy}_{\bar{c}}, \hat{\rvy}_{\bar{c}})$ + $\beta$ ${\Tilde{O}}(\hat{\rvy}_{\bar{c}})$  \Comment{\cref{eq5}}\;
        $\mathbf{final\_loss}$ $\gets$ $\mathcal{L}_{\text{COL}}$ + $\mathcal{Z}_{\text{VNE}} $  \Comment{\cref{eq:final}}\;
        optimizer.step($\mathbf{final\_loss}$)\;
 }
 \caption{Training WeedCLR}
 \label{alg:pseudo_code}
\end{algorithm}

As shown in \cref{alg:pseudo_code}, this work combines the Uniform Prior Loss and the Optimized Loss with a single entropy during the training process (see line 8 in \cref{alg:pseudo_code}). To balance the Uniform Prior Loss, \cref{eq:L_sym}, and the Optimized Loss, \cref{eq:complement_cross_entropy}, a coefficient  $\beta$ is introduced to the Optimized Loss, as shown in   \cref{eq4}. 

\begin{equation}\label{eq4}
\begin{aligned}
    \beta = \frac{\mathbf{\gamma}}{K-1}
\end{aligned}
\end{equation}
where $\gamma$  is a modulating factor, and $K$ is the number of classes.

This equation calculates the coefficient  $\beta$ as the ratio of the modulating factor $\gamma$ to the total number of classes minus one ($K-1$). The modulating factor $\gamma$ should be tuned to decide the amount that optimizes the Uniform Prior Loss, for example, $\gamma=-1$ ($\gamma<0$).

The proposed loss, named Class-optimized Loss (COL), is defined in   \cref{eq5}. 

\begin{equation}\label{eq5}
\begin{aligned}
    \mathcal{L}_{\text{COL}} = \mathcal{H}({\rvy}_{\bar{c}}, \hat{\rvy}_{\bar{c}}) + \beta {\Tilde{O}}(\hat{\rvy}_{\bar{c}})
\end{aligned}
\end{equation}
This equation calculates $\mathcal{L}_{\text{COL}}$ as the sum of the Uniform Prior Loss $\mathcal{H}({\rvy}_{\bar{c}}, \hat{\rvy}_{\bar{c}})$ and the product of the coefficient  $\beta$ and the Optimized Loss ${\Tilde{O}}(\hat{\rvy}_{\bar{c}})$. 

The final loss is then calculated by combining COL \cref{eq5} and VNE loss \cref{equ:vne}, as shown in equation \cref{eq:final}.


\begin{equation}\label{eq:final}
\begin{aligned}
 \mathcal{L}_{\text{COL}} + \mathcal{Z}_{\text{VNE}} 
\end{aligned}
\end{equation}

This approach allows for explicit optimization of model behavior for both correct and incorrect classes while taking into account information from both. To visualize the effect of our Class-Optimized Loss (COL), \cref{fig:6} shows the embeddings for DeepWeeds and CottonWeedID15 \citep{Chen2022cotton} test images obtained from two self-supervised training paradigms. On the left, the embeddings were obtained by training with cross-entropy only, while on the right, the embeddings were obtained by training with our COL. The model used is ResNet-50, and the embeddings represent the vector representation before feeding to the MLP and classification heads. The embeddings were projected to 2D vectors using t-SNE for visualization purposes. Compared to the left images, the clusters of each class in the right images are narrower in terms of intra-cluster distance. Additionally, the clusters in the right images have clean and separable boundaries, leading to more accurate and robust classification results.

\section{Experiments} \label{secExperiments}
This section first briefly overviews the experimental setup and implementation details and then presents experimental results demonstrating the effectiveness of our proposed method for weed classification in long-tailed datasets. We conducted experiments on two datasets of weed images, comparing the performance of our method to several baseline approaches.

\begin{figure*}[!t]
\centering
\includegraphics[width=0.98\textwidth]{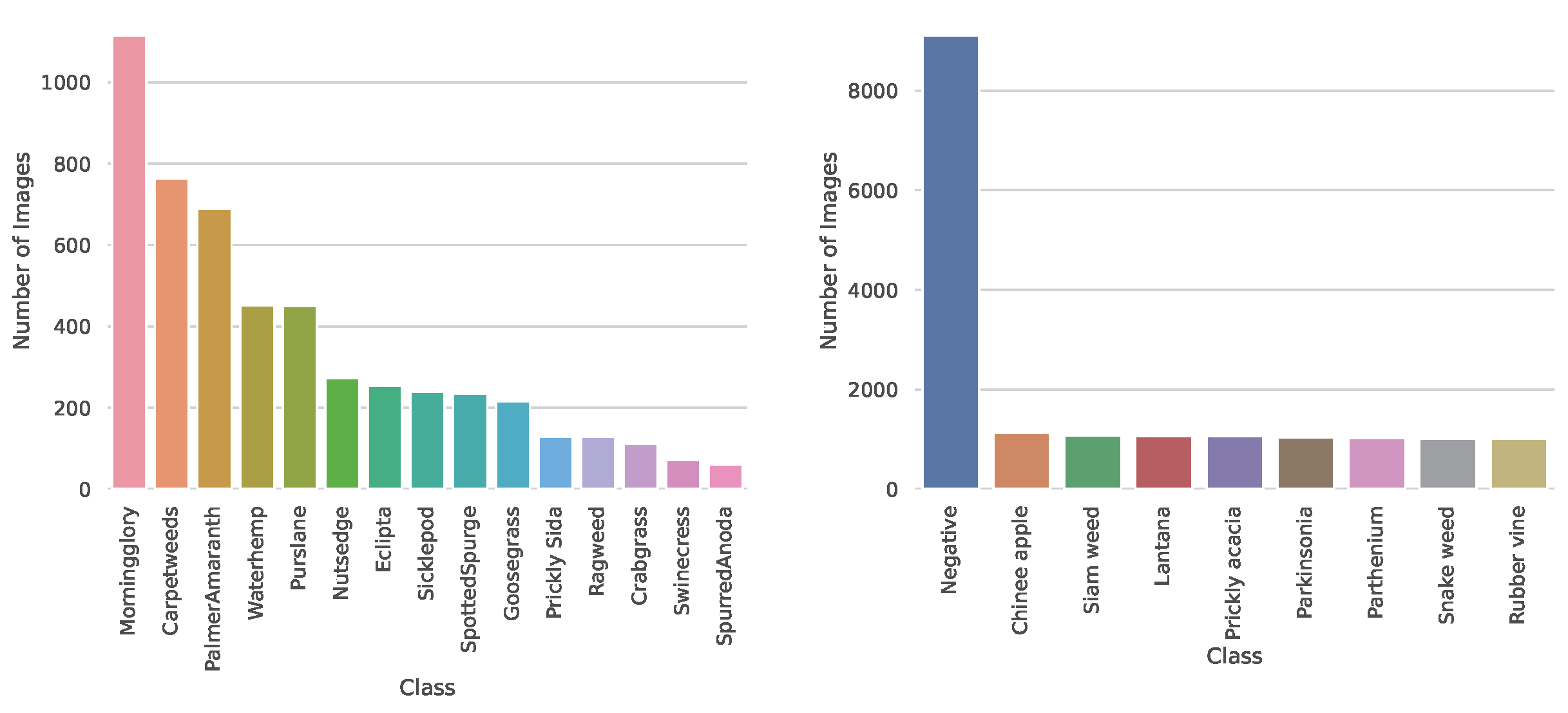}
\caption{Comparison of the number of images per class for two imbalanced datasets: CottonWeedID15  \citep{Chen2022cotton} (left) and DeepWeeds \citep{Olsen2019DeepWeed}  (right).}
\label{fig:1}
\end{figure*}

\subsection{Datasets} 
The first dataset used in our experiments is the DeepWeeds dataset \citep{Olsen2019DeepWeed}, which consists of 17,509 images capturing eight different weed species native to Australia in situ with neighbouring flora. The images were collected from weed infestations in eight rangeland environments across northern Australia. The second dataset is the CottonWeedID15 dataset \citep{Chen2022cotton}, which consists of 5,187 RGB images of 15 weeds that are common in cotton fields in the southern U.S. states. These images were acquired by either smartphones or hand-held digital cameras, under natural field light conditions and at varied stages of weed growth in 2020 and 2021. 
\cref{fig:1} compares the number of images per class for the two imbalanced datasets, CottonWeedID15 (left) and DeepWeeds  (right). 
Both datasets have a long-tailed distribution, with some weed species being much more common than others. These datasets provide a valuable resource for the development and evaluation of self-supervised learning methods for weed identification.

\subsection{Architecture}
In our experiments, we employed a ResNet-50 architecture, a commonly utilized backbone in self-supervised learning studies.
We also experimented with ResNet-9 and ResNet-18 architectures.
The ResNet-(9, 18, 50) backbones were initialized randomly. The projection heads used in the experiments were two-layer MLPs with sizes of $4096$ and one-layer with sizes of $128$, respectively, and included batch normalization, leaky-ReLU activations, and $\ell_2$ normalization after the last layer. Four classification heads were placed on top of the projection head MLPs, corresponding to $0.5K$, $1K$, $1.5K$, and $2K$ classes, respectively. Each classification head was a simple linear layer without an additive bias term. The row-softmax temperature $\tau_{row}$ was set to $0.1$, while the column-softmax temperature $\tau_{col}$ was set to $0.05$. The evaluation for unsupervised classification was conducted using two different classification heads. For linear probe evaluation, the MLP was removed and replaced with a single linear layer of $15$ and $9$ classes for CottonWeedID and DeepWeeds datasets, respectively.
 A schematic diagram of the architecture is depicted in \cref{fig:3}.

\subsection{Image Augmentations}
\begin{figure*}[t]
\centering
\includegraphics[width=0.98\textwidth]{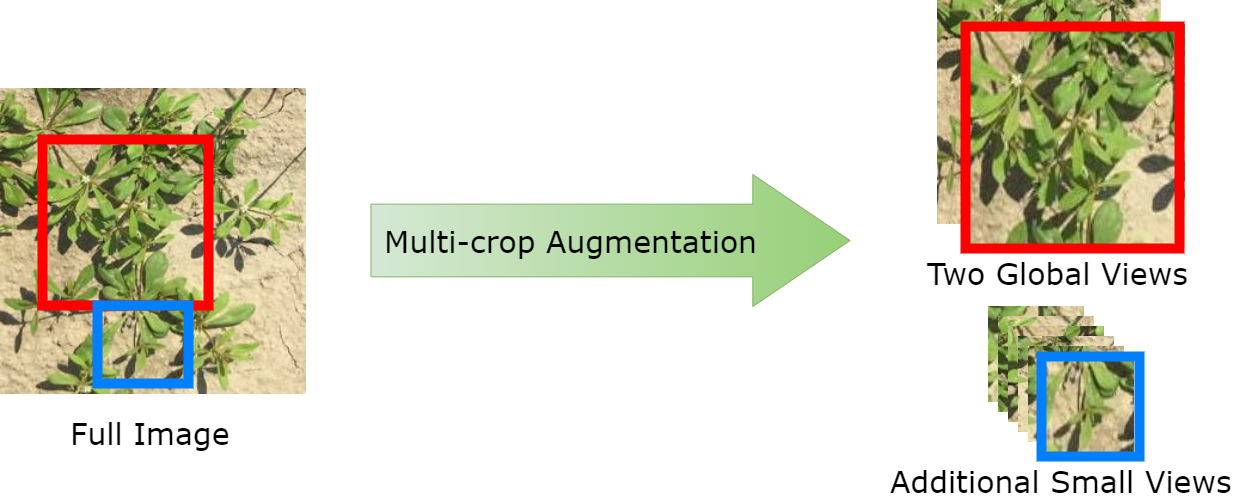}
\caption{\textbf{Multi-Crop Augmentation}:  This process transforms a single image into V+2 distinct views, comprising two global perspectives and V small resolution zoomed perspectives. By doing so, it introduces a higher level of diversity into the training data, thereby enhancing the robustness and generalization ability of the trained model.}
\label{fig:10}
\end{figure*}

In our experiments, we employed the data augmentations of SimCLR \citep{Chen2020b}, including colour jittering, Gaussian blur, and random flips. We also utilized a multi-crop strategy, with two global views of size $128\times128$ and four local views of size $64\times64$, as well as nearest neighbour augmentation, with a queue set to $200$. The multi-crop strategy involves creating multiple crops of different sizes from the same image for training a model, see \cref{fig:10}. This approach was introduced in the SwAV \citep{Caron2020UnsupervisedAssignments} method for unsupervised learning of visual features. In SwAV, instead of using a fixed size for cropping images, a multi-crop strategy is employed where two larger crops and up to four smaller crops are taken from the same image. This has been shown to boost the performance of the model compared to other approaches that use a fixed crop size. The proposed strategy is simple, yet effective, and can be applied to many self-supervised methods to consistently improve performance.

\subsection{Experimental Setup} 

For unsupervised pre-training and classification, we  used an SGD optimizer \citep{you2017large} with a learning rate of $4.8$ and weight decay of $10^{-6}$. The learning rate was linearly ramped up from $0.3$ over the first $10$ epochs, and then decreased using a cosine scheduler for $390$ epochs with a final value of $0.0048$, for a total of $400$ epochs. We used a batch size of $256$ on a single NVIDIA GeForce RTX 2080 Ti GPU.

We measured the quality of our WeedCLR model representations using two approaches. 
The first approach involves using the k-nearest neighbours (KNN) classifier, which makes classifications based on proximity to other data points.
The second approach,  known as linear probing, involves using the pre-trained model as a feature extractor. Given labelled examples ($X, Y$), the model is applied to $X$ to produce features $f_X$. A linear classifier is then trained on ($f_X, Y$). Linear probing captures the intuition that good features should linearly separate the classes of transfer tasks and helps disentangle feature quality from model architecture.

For the linear probe evaluation, we used a similar experimental approach to KNN and used an SGD optimizer \citep{you2017large} with a learning rate of $0.1$ and no weight decay. The learning rate was decreased using a cosine scheduler for $100$ epochs. We used a batch size of $256$ on a single NVIDIA GeForce RTX 2080 Ti GPU.

\begin{figure*}[t]
\centering
\includegraphics[width=0.98\textwidth]{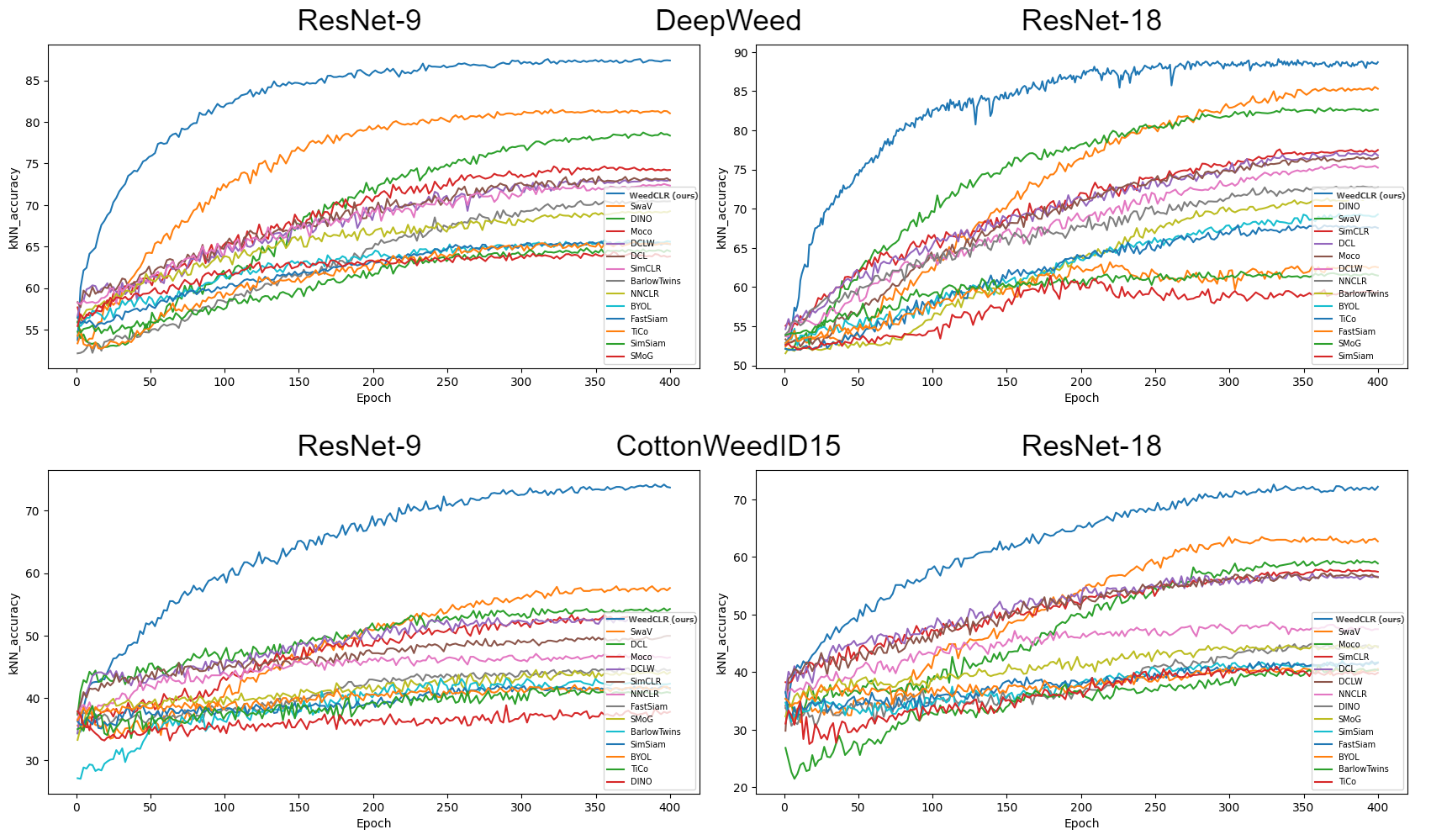}
\caption{Performance comparison of our WeedCLR method with other state-of-the-art self-supervised learning methods on the DeepWeeds and CottonWeedID15 datasets. Our results show that WeedCLR learns faster, reaching 76\% online KNN accuracy in just 50 epochs, while SwaV \citep{Caron2020UnsupervisedAssignments} requires 200 epochs to achieve the same level of accuracy. With an increased ResNet from 9 to 18 and more epochs, WeedCLR achieves 89\% accuracy on DeepWeeds and 73\% accuracy using a small batch size, a small number of stored features, and no momentum encoder. These results demonstrate the effectiveness of our approach in improving classification accuracy compared to state-of-the-art self-supervised learning methods. \textit{The figure may be better viewed online where you can zoom in for more detail.}}
\label{fig:2}
\end{figure*}

\subsection{Online Training Accuracy}
In this section, we compare the performance of our WeedCLR method with other state-of-the-art self-supervised learning methods on the DeepWeeds and CottonWeedID15 datasets during training. \cref{fig:2} shows that WeedCLR learns faster, reaching higher online KNN accuracy in fewer epochs than the other methods. With an increased ResNet from 9 to 18 and more epochs, WeedCLR still achieves high accuracy and maintains its advantage over the other methods. These results demonstrate the effectiveness of our approach in improving classification accuracy compared to state-of-the-art self-supervised learning methods.
The following section quantitatively compares our method with other state-of-the-art methods in more detail.

\section{Results} \label{ssecrslt}

In some studies,  an alternative approach to our utilized linear probe evaluation or KNN may be employed to measure the quality of representations. This involves fine-tuning the pre-trained model for image classification tasks. The process includes adding a small classification head to the model and adjusting all weights accordingly. 
However, it is important to note that during fine-tuning, one model may outperform another not necessarily because of superior pre-training, but possibly because its architecture is more compatible with the downstream task at hand. Therefore, we did not use fine-tuning as a representation quality metric in our study.
Accordingly, our approach consists of a pre-training stage followed by a linear probe and a KNN classifier. In pre-training, we explore three variants of the ResNet architecture (ResNet-9, 18, 50) as a backbone.

\begin{figure*}[t]
\centering
\includegraphics[width=0.98\textwidth]{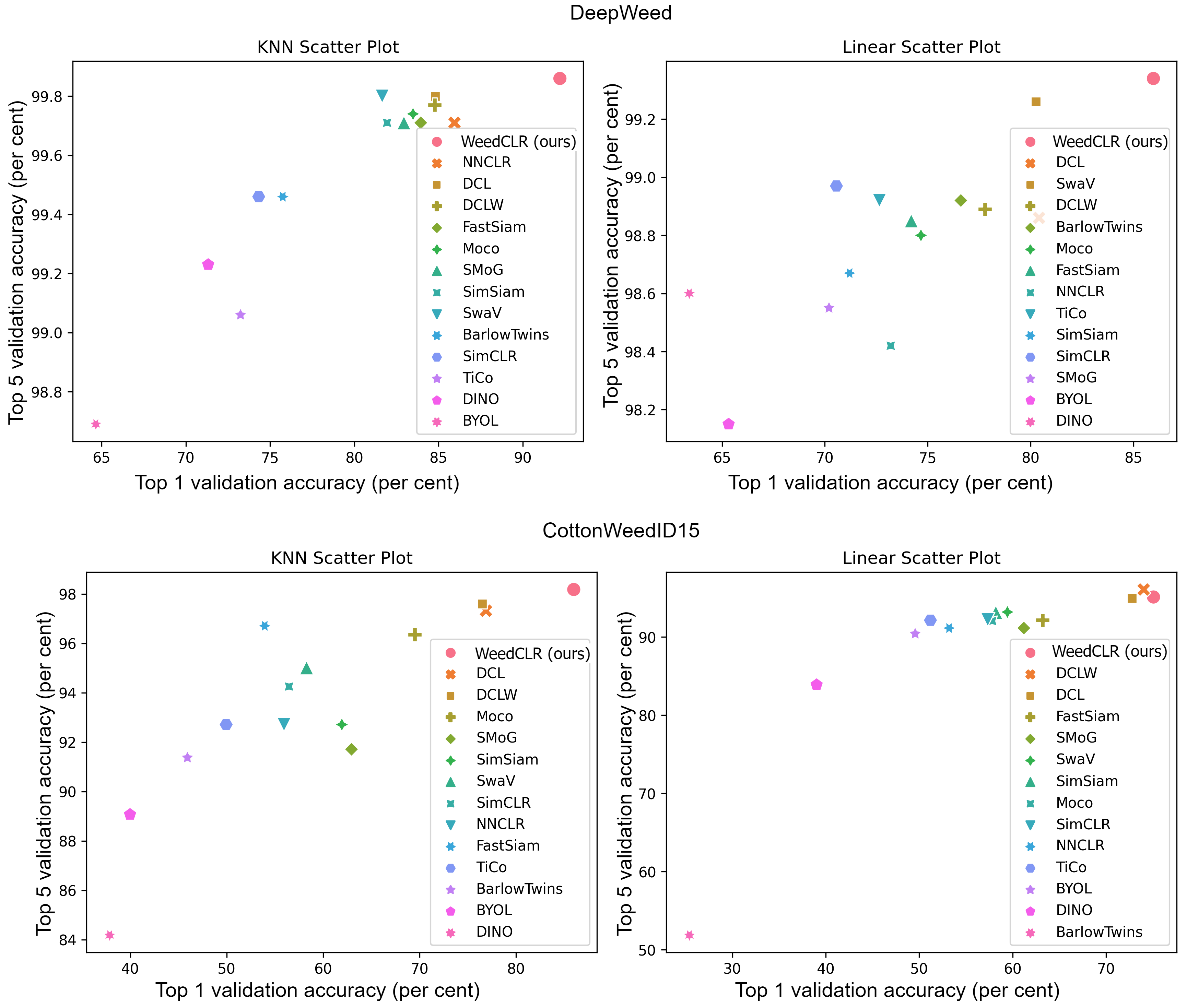}
\caption{Performance comparison of our approach, WeedCLR, against other self-supervised learning methods using both KNN classifier and linear probe evaluations on the DeepWeeds and CottonWeedID datasets. Our results demonstrate that WeedCLR outperformed other methods in terms of top-1 and top-5 accuracy for both datasets and accuracy metrics. For more details, please see Tables \ref{tab:DWknn}, \ref{tab:DWLinear}, \ref{tab:c15knn}, and \ref{tab:c15Linear}. The figure may be better viewed online where you can zoom in for more detail.}
\label{fig:5}
\end{figure*}

In our experiments, similar to the online training accuracy test we performed, we evaluated the performance of our approach, WeedCLR, against other self-supervised learning methods using both 
KNN classifier and linear probe evaluations on the DeepWeeds and CottonWeedID datasets. Our results, presented in Tables \ref{tab:DWknn}, \ref{tab:DWLinear}, \ref{tab:c15knn}, and \ref{tab:c15Linear}, demonstrate that WeedCLR outperformed other methods in terms of top-1 and top-5 accuracy for all three variants of the ResNet architecture (ResNet-9,18,50).
Moreover, \cref{fig:5} compares the performance of our approach, WeedCLR, against other self-supervised learning methods using both KNN classifier and linear probe evaluations using ResNet-50.

\subsection{DeepWeeds dataset}

The results of the KNN classification and linear probing experiments are shown in Tables \ref{tab:DWknn} and \ref{tab:DWLinear}, respectively. WeedCLR achieved the best top-1 and top-5 accuracies for all three ResNet architectures on both evaluation metrics.  
Our method, WeedCLR, outperformed the second-best model across all metrics and ResNet architectures. This superiority can be attributed to WeedCLR's ability to learn more discriminative features, which is crucial for accurate classification.

When compared to the supervised model, WeedCLR shows competitive performance. This is particularly noteworthy given that WeedCLR is a self-supervised method, which typically has a harder task as it does not have access to label information during training.

Interestingly, the top-5 accuracy of WeedCLR is very close to that of the supervised model. This suggests that while the top-1 predictions might differ, the set of top-5 predictions between the two models are quite similar. This could be due to both models learning similar feature representations for the classes, leading to similar predictions among the top-5 classes.

In conclusion, these results demonstrate the effectiveness of WeedCLR in learning discriminative features for weed classification, even in a self-supervised setting.

\subsection{CottonWeedID dataset}

The results of the KNN classification and linear probing experiments are shown in Tables \ref{tab:c15knn} and \ref{tab:c15Linear}, respectively. 
WeedCLR consistently outperforms other methods in most of the evaluations, demonstrating its robustness and effectiveness. It is particularly noteworthy that WeedCLR achieves the best top-1 and top-5 accuracy across all three backbones using the KNN classifier. This suggests that the features learned by WeedCLR are highly discriminative, enabling accurate nearest neighbor classification.

In the linear probing evaluation, WeedCLR continues to show strong performance, achieving the best results in 5 out of 6 tests. The only exception is for ResNet-9 backbone for Top-5 accuracy, where WeedCLR comes second. This could be due to the lower capacity of ResNet-9 compared to the other architectures, which might limit the effectiveness of the learned representations.

Overall, the results of our experiments demonstrate that WeedCLR is a powerful self-supervised learning method for plant classification. Our approach can be used to learn high-quality representations that are capable of distinguishing between different plant species. In addition to our quantitative results, we also conducted qualitative experiments in the next section to visualize the feature representations learned by WeedCLR. These experiments showed that WeedCLR learned discriminative features that can effectively separate different plant species.



 \begin{table}[h]
 \caption{
k-nearest neighbours classifier evaluation on DeepWeeds  \citep{Olsen2019DeepWeed} 
}
\label{tab:DWknn}
\centering
\resizebox{0.98\columnwidth}{!}{
\begin{tabular}{@{}lcccccc@{}}
\toprule
Method    & \multicolumn{2}{c}{ResNet-9}     & \multicolumn{2}{c}{ResNet-18}  & \multicolumn{2}{c}{ResNet-50}     \\ 
\cmidrule(lr){2-3} \cmidrule(lr){4-5} \cmidrule(lr){6-7}
                & Top-1         & Top-5       & Top-1         & Top-5  & Top-1         & Top-5        \\ 
                                                      \midrule
Supervised \citep{Olsen2019DeepWeed}     & 97.51     &99.92         & 97.64     &99.93      &97.34    &99.92  \\ 
\midrule
       BYOL \citep{grill2020bootstrap}               & 64.66     &98.69         & 69.94     &99.03      &69.08    &98.50   \\
BarlowTwins \citep{Zbontar2021BarlowReduction}       & 75.76     &99.46         & 75.39     &99.34      &74.72    &98.41   \\ 
        DCL \citep{Yeh2022DecoupledLearning}         & 84.80     &99.80         & 87.02     &99.89      &86.39    &99.04    \\
       DCLW \citep{Yeh2022DecoupledLearning}         & 84.77     &99.77         & 86.42     &99.77      &85.73    &98.78    \\ 
       DINO \citep{Caron2021EmergingTransformers}    & 71.33     &99.23         & 78.58     &99.49      &77.89    &98.68    \\ 
   FastSiam \citep{Pototzky2022FastSiam:GPU}         & 83.95     &99.71         & 83.30     &98.79      &82.78    &97.79    \\ 
       Moco \citep{Chen2020ImprovedLearning}         & 83.49     &99.74         & 84.57     &99.74      &83.63    &99.03    \\ 
      NNCLR \citep{Dwibedi2021WithRepresentations}   & 85.95     &99.71         & 85.13     &99.14      &84.36    &98.28    \\ 
       SMoG \citep{Pang2022UnsupervisedGrouping}     & 82.95     &99.71         & 82.31     &98.82      &81.38    &98.06    \\ 
     SimCLR \citep{Chen2020b}                        & 74.33     &99.46         & 78.64     &99.60      &77.80    &99.00    \\ 
    SimSiam \citep{Xie2022SimMIM:Modeling}           & 81.95     &99.71         & 81.03     &98.83      &80.36    &98.32    \\ 
       SwaV \citep{Caron2020UnsupervisedAssignments} & 81.66     &99.80         & 83.37     &99.74      &82.49    &98.82    \\ 
       TiCo \citep{Zhu2022TiCo:Learning}             & 73.25     &99.06         & 76.95     &99.34      &76.45    &98.52    \\ 
\midrule
WeedCLR (ours)  & \textbf{92.21}     &\textbf{99.86}         & \textbf{91.93}     &\textbf{99.77 }     &\textbf{91.21}    &\textbf{99.76}   \\

\bottomrule
\end{tabular}
}
\vspace{-0.2cm}

\end{table}

\begin{table}[h]
\caption{
Linear probe evaluation on DeepWeeds  \citep{Olsen2019DeepWeed} 
}
\label{tab:DWLinear}
\centering
\resizebox{0.98\columnwidth}{!}{
\begin{tabular}{@{}lcccccc@{}}
\toprule
Method    & \multicolumn{2}{c}{ResNet-9}     & \multicolumn{2}{c}{ResNet-18}  & \multicolumn{2}{c}{ResNet-50}     \\ 
\cmidrule(lr){2-3} \cmidrule(lr){4-5} \cmidrule(lr){6-7}
                & Top-1         & Top-5       & Top-1         & Top-5  & Top-1         & Top-5        \\ 
                                                      \midrule
Supervised  \citep{Olsen2019DeepWeed}     & 94.17          & 99.73         & 94.84          & 99.83    &  95.70 & 99.33  \\ 
\midrule
       BYOL \citep{grill2020bootstrap}                   &65.32     &98.15          &67.17     &98.60    &66.59   &97.97   \\
BarlowTwins \citep{Zbontar2021BarlowReduction}           &76.61     &98.92          &74.81     &99.00    &73.89   &98.03    \\ 
        DCL \citep{Yeh2022DecoupledLearning}             &80.41     &98.86          &79.89     &98.49    &78.89   &97.79   \\
       DCLW \citep{Yeh2022DecoupledLearning}             &77.78     &98.89          &81.72     &99.26    &81.11   &98.38   \\ 
       DINO \citep{Caron2021EmergingTransformers}        &63.41     &98.60          &75.07     &98.83    &74.14   &98.29   \\ 
   FastSiam \citep{Pototzky2022FastSiam:GPU}             &74.20     &98.85          &73.58     &98.19    &72.91   &97.23   \\ 
       Moco \citep{Chen2020ImprovedLearning}             &74.67     &98.80          &76.70     &99.14    &76.01   &98.43   \\ 
      NNCLR \citep{Dwibedi2021WithRepresentations}       &73.20     &98.42          &72.69     &97.71    &71.93   &97.03   \\ 
       SMoG \citep{Pang2022UnsupervisedGrouping}         &70.20     &98.55          &69.34     &97.65    &68.38   &97.05   \\ 
     SimCLR \citep{Chen2020b}                            &70.56     &98.97          &76.70     &99.29    &76.03   &98.46   \\ 
    SimSiam \citep{Xie2022SimMIM:Modeling}               &71.20     &98.67          &70.22     &98.17    &69.32   &97.32   \\ 
       SwaV \citep{Caron2020UnsupervisedAssignments}     &80.26     &99.26          &80.98     &99.17    &80.38   &98.56   \\ 
       TiCo \citep{Zhu2022TiCo:Learning}                 &72.65     &98.92          &74.33     &99.03    &73.51   &98.06   \\ 
\midrule
WeedCLR (ours)  &\textbf{85.97}     &\textbf{99.34}          &\textbf{87.22 }    &\textbf{99.49 }   &\textbf{86.63}   &\textbf{99.33}  \\ 
\bottomrule
\end{tabular}
}
\vspace{-0.2cm}

\end{table}


\begin{table}[h]
\caption{
k-nearest neighbours classifier evaluation on CottonWeedID  \citep{Chen2022cotton}
}
\label{tab:c15knn}
\centering
\resizebox{0.98\columnwidth}{!}{
\begin{tabular}{@{}lcccccc@{}}
\toprule
Method    & \multicolumn{2}{c}{ResNet-9}     & \multicolumn{2}{c}{ResNet-18}  & \multicolumn{2}{c}{ResNet-50}     \\ 
\cmidrule(lr){2-3} \cmidrule(lr){4-5} \cmidrule(lr){6-7}
                & Top-1         & Top-5       & Top-1         & Top-5  & Top-1         & Top-5        \\ 
                                                      \midrule
Supervised \citep{Chen2022cotton}     & 96.14     &99.15         &  97.32     &99.68      &98.01    &99.21  \\ 
\midrule
       BYOL \citep{grill2020bootstrap}                   & 39.98     &89.07         &  40.94     &90.12      &40.05    &89.55   \\
BarlowTwins \citep{Zbontar2021BarlowReduction}           & 45.93     &91.37         &  42.19     &90.03      &41.57    &89.31   \\ 
        DCL \citep{Yeh2022DecoupledLearning}             & 76.89     &97.32         &  79.67     &97.03      &78.88    &96.27    \\
       DCLW \citep{Yeh2022DecoupledLearning}             & 76.51     &97.60         &  77.66     &96.84      &76.98    &96.20    \\ 
       DINO \citep{Caron2021EmergingTransformers}        & 37.87     &84.18         &  41.51     &87.15      &40.78    &86.50    \\ 
   FastSiam \citep{Pototzky2022FastSiam:GPU}             & 53.95     &96.71         &  53.15     &95.96      &52.31    &95.26    \\ 
       Moco \citep{Chen2020ImprovedLearning}             & 69.51     &96.36         &  72.29     &96.93      &71.54    &96.02    \\ 
      NNCLR \citep{Dwibedi2021WithRepresentations}       & 55.95     &92.71         &  55.10     &91.97      &54.36    &91.33    \\ 
       SMoG \citep{Pang2022UnsupervisedGrouping}         & 62.95     &91.71         &  62.20     &90.81      &61.23    &90.19    \\ 
     SimCLR \citep{Chen2020b}                            & 56.47     &94.25         &  63.18     &96.07      &62.35    &95.09    \\ 
    SimSiam \citep{Xie2022SimMIM:Modeling}               & 61.95     &92.71         &  61.30     &91.82      &60.64    &90.89    \\ 
       SwaV \citep{Caron2020UnsupervisedAssignments}     & 58.29     &95.01         &  65.77     &95.88      &65.20    &95.04    \\ 
       TiCo \citep{Zhu2022TiCo:Learning}                 & 49.95     &92.71         &  49.95     &91.18      &49.21    &90.21    \\ 
\midrule
WeedCLR (ours)  & \textbf{86.00 }    &\textbf{98.18}         &  \textbf{84.66}     &\textbf{98.47}      &\textbf{83.84}    &\textbf{97.52} \\ 
\bottomrule
\end{tabular}
}
\vspace{-0.2cm}

\end{table}

\begin{table}[h]
\caption{
Linear probe evaluation  on CottonWeedID   \citep{Chen2022cotton}
}
\label{tab:c15Linear}
\centering
\resizebox{0.98\columnwidth}{!}{
\begin{tabular}{@{}lcccccc@{}}
\toprule
Method    & \multicolumn{2}{c}{ResNet-9}     & \multicolumn{2}{c}{ResNet-18}  & \multicolumn{2}{c}{ResNet-50}     \\ 
\cmidrule(lr){2-3} \cmidrule(lr){4-5} \cmidrule(lr){6-7}
                & Top-1         & Top-5       & Top-1         & Top-5  & Top-1         & Top-5        \\ 
                                                      \midrule
Supervised  \citep{Chen2022cotton}    & 95.32     &99.15         &  96.31     &99.68      &98.14    &99.34 \\ 
\midrule
       BYOL \citep{grill2020bootstrap}                   & 49.57     &90.41          &  45.06     &90.22    &44.46    &89.53   \\
BarlowTwins \citep{Zbontar2021BarlowReduction}           & 25.41     &51.87          &  15.15     &46.02    &14.22    &45.13    \\ 
        DCL \citep{Yeh2022DecoupledLearning}             & 72.77     &94.97          &  74.22     &95.12    &74.03    &94.33   \\
       DCLW \citep{Yeh2022DecoupledLearning}             & 74.02     &\textbf{96.07}          &  73.06     &95.45    &72.35    &94.26   \\ 
       DINO \citep{Caron2021EmergingTransformers}        & 39.02     &83.89          &  43.43     &86.67    &42.45    &85.78   \\ 
   FastSiam \citep{Pototzky2022FastSiam:GPU}             & 63.20     &92.14          &  62.26     &91.46    &61.72    &90.90   \\ 
       Moco \citep{Chen2020ImprovedLearning}             & 57.81     &92.04          &  61.27     &93.48    &60.42    &92.51   \\ 
      NNCLR \citep{Dwibedi2021WithRepresentations}       & 53.20     &91.14          &  52.24     &90.42    &51.58    &89.60   \\ 
       SMoG \citep{Pang2022UnsupervisedGrouping}         & 61.20     &91.14          &  60.37     &90.52    &59.44    &89.96   \\ 
     SimCLR \citep{Chen2020b}                            & 57.33     &92.23          &  63.95     &94.82    &63.25    &93.94   \\ 
    SimSiam \citep{Xie2022SimMIM:Modeling}               & 58.20     &93.14          &  57.24     &92.32    &56.47    &91.44   \\ 
       SwaV \citep{Caron2020UnsupervisedAssignments}     & 59.44     &93.19          &  66.25     &94.63    &65.30    &94.10   \\ 
       TiCo \citep{Zhu2022TiCo:Learning}                 & 51.20     &92.14          &  48.42     &90.80    &47.47    &90.23   \\ 
\midrule
WeedCLR (ours) & \textbf{75.07 }    &95.11          &  \textbf{75.55}     &\textbf{95.59}    &\textbf{75.04 }   &\textbf{94.64} \\ 
\bottomrule
\end{tabular}
}
\vspace{-0.2cm}

\end{table}

\clearpage

\subsection{Qualitative Results}
\begin{figure*}[!t]
\centering
\includegraphics[width=0.80\textwidth]{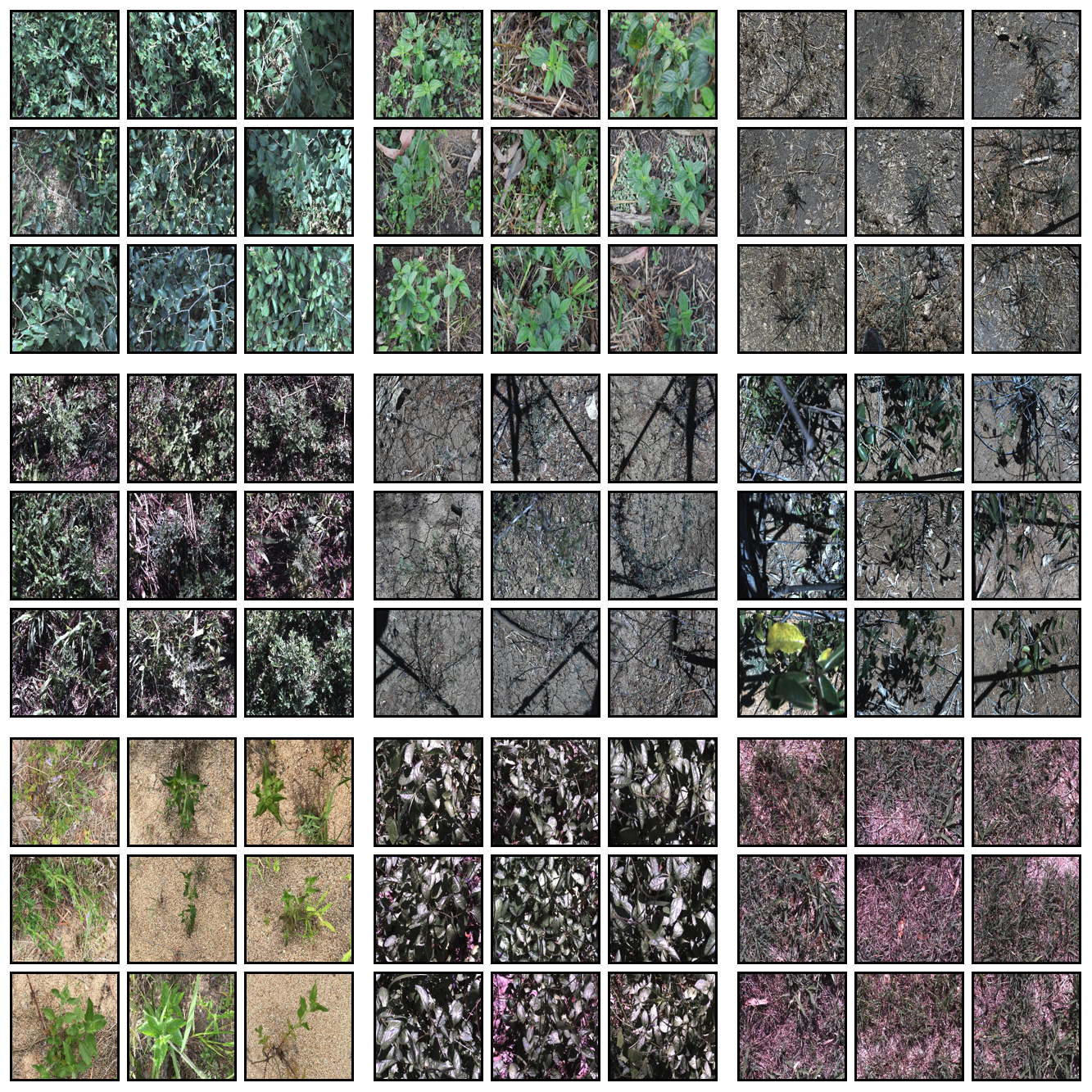}
\caption{ Classes predicted with high accuracy by our \textit{WeedCLR} on the DeepWeeds  \citep{Olsen2019DeepWeed} validation set, which was not seen during training. The images shown are randomly selected from each predicted class, namely from the top left:  Chinee apple,  Lantana,  Parkinsonia,   Parthenium,   Prickly acacia,  Rubber vine,   Siam weed,   Snake weed and  Negatives. The variety of backgrounds and weeds in the predicted classes demonstrates that the \textit{WeedCLR}  is able to learn semantically meaningful classes without the use of labels.}
\label{fig:7}
\end{figure*}

\begin{figure*}[!t]
\centering
\includegraphics[width=0.65\textwidth]{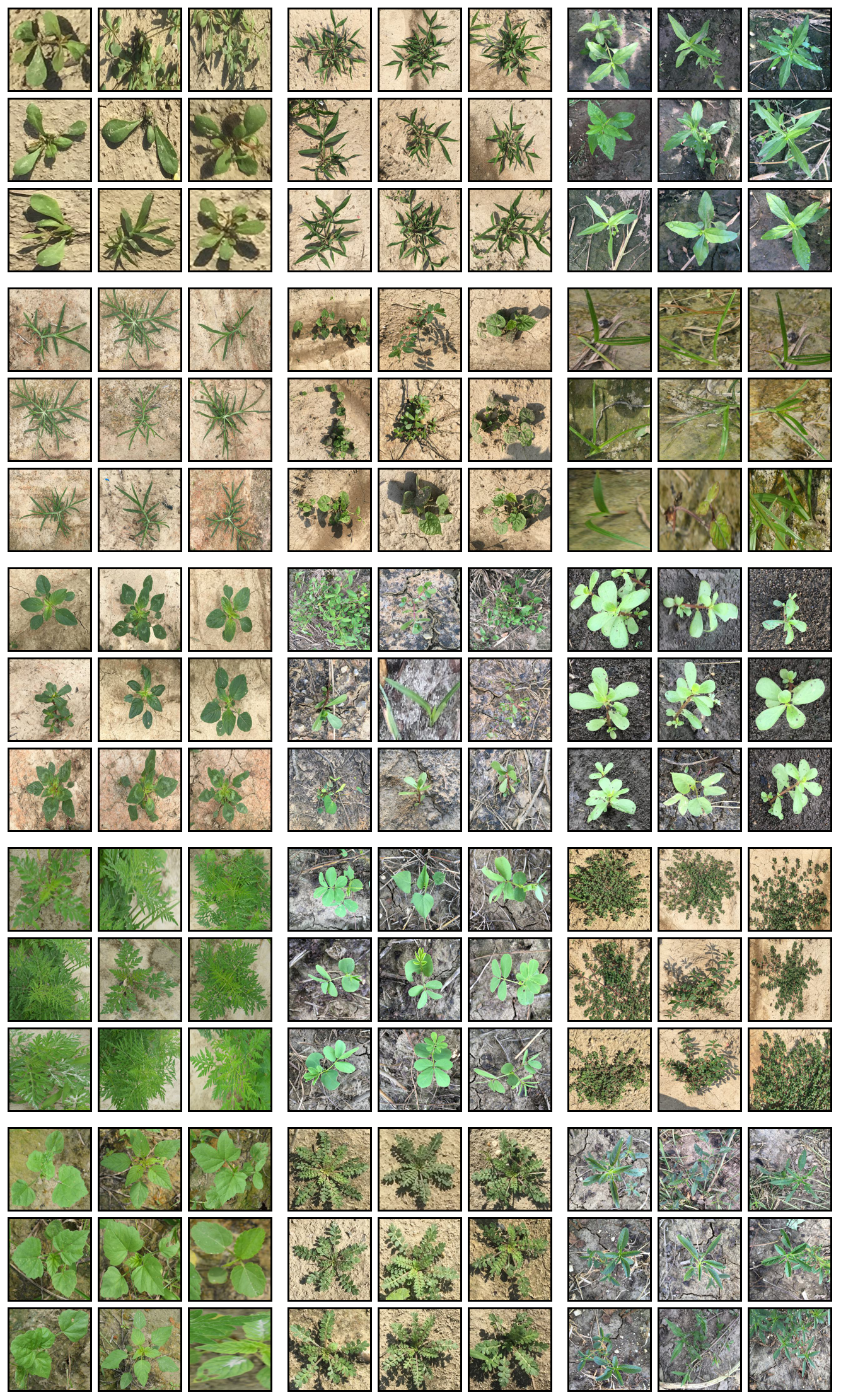}
\caption{ Classes predicted with high accuracy by our \textit{WeedCLR} on the CottonWeedID15  \citep{Chen2022cotton} validation set. The images shown are randomly selected from each predicted class, namely from the top left:  Carpetweeds, Crabgrass, Eclipta, Goosegrass, Morningglory, Nutsedge, PalmerAmaranth, Prickly Sida, Purslane, Ragweed, Sicklepod, SpottedSpurge, SpurredAnoda, Swinecress, Waterhemp. }
\label{fig:8}
\end{figure*}

In this section, we present qualitative results from our experiments with WeedCLR. These results provide a visual representation of the effectiveness of our approach in learning semantically meaningful classes without the use of labels. \cref{fig:7} and \cref{fig:8} show sample images of classes predicted with high accuracy by our WeedCLR method on the DeepWeeds \citep{Olsen2019DeepWeed} and CottonWeedID15 \citep{Chen2022cotton} validation sets, respectively. The images shown are randomly selected from each predicted class. The variety of backgrounds and weeds in the predicted classes demonstrates the ability of our WeedCLR method to learn high-quality representations for plant classification.

\section{Ablation Study} \label{secabl}

This ablation study investigates the effect of different factors on the top-1 accuracy of KNN and linear classifiers on the DeepWeeds and CottonWeedID datasets. The factors considered are the loss function generality, batch size, softmax temperature, and MLP architecture.

\subsection{Loss Function Generality}

The loss function generality refers to the ability of the loss function to handle imbalanced datasets. The baseline model uses the cross-entropy loss function, which is not as robust to imbalanced datasets as other loss functions. The VNE is optimizing the representation space that can achieve a more desirable representation that avoids dimensional collapse and produces more useful embeddings.    The COL loss function is our proposed variant of the cross-entropy loss function that is more robust to imbalanced datasets.

The results shown in Table \ref{table:ablation_Parameter} demonstrate that the VNE and COL loss functions consistently and significantly improve the top-1 accuracy of the KNN and linear classifiers on both datasets. The (VNE + COL) loss function achieves the best results, with a top-1 accuracy of 91.21\% on the DeepWeeds dataset and 83.84\% on the CottonWeedID dataset.

\begin{table}[h] 
\centering
\caption{Loss function generality - Top-1 accuracy}
\label{table:ablation_Parameter}
\begin{tabular}{l|cc|cc}
\toprule
 & \multicolumn{2}{c|}{DeepWeed}& \multicolumn{2}{c}{ CottonWeedID} \\
Parameter            & KNN   & Linear   & KNN   & Linear   \\
\midrule
Baseline             & 54.35 & 55.13 & 45.78 & 45.42 \\
Baseline + COL       & 84.76 & 82.84 & 72.63 & 67.89 \\
Baseline + VNE       & 87.68 & 84.32 & 79.51 & 71.48 \\
Baseline + COL + VNE & 91.21 & 86.63 & 83.84 & 75.04 \\
\bottomrule
\end{tabular}
\end{table}

\subsection{Batch Size}

The batch size refers to the number of samples that are processed in each iteration of the training algorithm. The baseline batch size is 32. The other batch sizes are 64, 128, and 256.

The results shown in Table \ref{table:ablation_batch_size} demonstrate that the accuracy of the top-1 of the KNN and linear classifiers increases with the batch size. It is clear that the most significant improvement in top-1 accuracy for both KNN and linear classifiers occurs when the batch size is increased from 128 to 256. This is true for both the DeepWeeds and CottonWeedID datasets.

This substantial improvement could be attributed to the fact that a larger batch size allows the model to estimate the gradient more accurately during training. However, it is also important to note that using a larger batch size requires more memory, which might not always be feasible depending on the hardware constraints.

In conclusion, while our method shows robust performance across different batch sizes, using a larger batch size of 256 leads to the best performance in terms of top-1 accuracy. Future work could explore the impact of even larger batch sizes on performance, given sufficient computational resources.

\begin{table}[h] 
\centering
\caption{Batch size - Top-1 accuracy}
\label{table:ablation_batch_size}
\begin{tabular}{c|cc|cc}
\toprule
 & \multicolumn{2}{c|}{DeepWeed}& \multicolumn{2}{c}{ CottonWeedID} \\
Batch Size & KNN   & Linear   & KNN   & Linear   \\
\midrule
32  & 82.24 & 77.31 & 76.82 & 70.34 \\
64  & 85.17 & 79.48 & 77.92 & 70.27 \\
128 & 87.34 & 80.27 & 79.61 & 72.98 \\
256 & 91.21 & 86.63 & 83.84 & 75.04 \\
\bottomrule
\end{tabular}
\end{table}

\subsection{Softmax Temperature}

The softmax temperature is a hyperparameter that controls the degree of confidence in the predicted probabilities assigned to each class. A higher temperature results in a softer probability distribution, where the predicted probabilities are more evenly distributed among the classes, while a lower temperature results in a sharper probability distribution, where the predicted probabilities are more concentrated on the most likely class.

The results shown in Table \ref{table:ablation_sm_temp} demonstrate  that the top-1 accuracy of the KNN classifier increases with the softmax temperature. The most significant improvement in top-1 accuracy for the KNN classifier occurs when the softmax temperature is increased from 0.03 to 0.05. This is true for both the DeepWeeds and CottonWeedID datasets. These results suggest that a higher softmax temperature can lead to improved performance of the KNN classifier in this context.

\begin{table}[h] 
\centering
\caption{Softmax Temperature - KNN classifier Top-1 accuracy}
\label{table:ablation_sm_temp}
\begin{tabular}{c|cc|cc}
\toprule
 & \multicolumn{4}{c}{$\tau_{row}$} \\
 \cmidrule(lr){3-4}
 & \multicolumn{2}{c|}{DeepWeed}& \multicolumn{2}{c}{CottonWeedID} \\
$\tau_{column}$ & 0.07 & 0.1& 0.07 & 0.1 \\
\midrule
0.03  & 90.78 & 91.04 & 82.45 & 82.78 \\
0.05  & 91.08 & 91.21 & 83.25 & 83.84 \\
\bottomrule
\end{tabular}
\end{table}

\subsection{MLP Architecture}

The MLP architecture refers to the number of hidden layers and the number of units in each hidden layer. The baseline MLP architecture has one hidden layer with 4096 units. The other MLP architectures have two hidden layers with 4096 units each and two hidden layers with 8192 units each.

The results shown in Table \ref{table:ablation_mlp} demonstrate  that the architecture with two hidden layers of 4096 units each provides slightly higher top-1 accuracy for the KNN classifier compared to the baseline architecture (one hidden layer with 4096 units). This suggests that adding an additional layer might help the model capture more complex patterns in the data, leading to improved performance.

However, it is important to note that increasing the size of the hidden layers (from 4096 to 8192 units) in a two-layer architecture does not lead to further improvements. This could be due to overfitting, where the model becomes too complex and starts to fit the noise in the training data rather than the underlying patterns.
In our results presented above, we have used the architecture with two hidden layers of 4096 units each, as it provided the best performance in the ablation study.  

\begin{table}[h] 
\centering
\caption{MLP architecture - KNN classifier  Top-1 accuracy}
\label{table:ablation_mlp}
\begin{tabular}{c|cc|cc}
\toprule
 & \multicolumn{4}{c}{MLP output layer  } \\
  \cmidrule(lr){3-4}
 & \multicolumn{2}{c|}{DeepWeed}& \multicolumn{2}{c}{CottonWeedID} \\
MLP hidden layer(s) & 128 & 256& 128 & 256 \\
\midrule
1x4096 & 90.78 & 89.01 & 82.27 & 81.65 \\
2x4096 & 91.21 & 90.34 & 83.84 & 82.61 \\
2x8192 & 90.32 & 88.98 & 82.78 & 81.64 \\
\bottomrule
\end{tabular}
\end{table}


\section{Discussion} \label{secdisc}
In this paper, we presented a novel method for weed classification in long-tailed datasets using a self-visual features learning approach. Our method, called Weed Contrastive Learning through visual Representation (WeedCLR), utilizes a class-optimized loss function to improve classification accuracy. We demonstrated the effectiveness of our approach on two datasets of weed images, achieving state-of-the-art performance.

Our results show that the proposed method is able to effectively learn discriminative visual features for weed classification without the use of any annotations even in the presence of long-tailed data distributions. This has significant implications for the development of automated weed management systems \citep{Arsa2023Eco-friendlyNetworks, Dang2023YOLOWeeds:Systems}. By removing the requirement of human labelling, this self-supervised approach could reduce the time and cost burden of existing fully supervised approaches. For example, \citep{calvert2021robotic} reports a time requirement of one hour to label 2,000 images for weed classification. For their 58,153 image dataset, approximately 29 hours of labelling by domain experts and the equivalent labour cost could be avoided, while achieving a similar or better classification accuracy. This increased efficiency also has a second advantage allowing for deep learning based approaches to be more rapidly deployed on-farm when targeting a new weed or crop scenario. Currently, the time it takes from dataset collection to field implementation is strongly influenced by the time to annotate images. With the removal of this annotation time, the window from dataset collection to implementation is shortened allowing for more timely and effective weed control on new target weed crop scenarios. As such, WeedCLR can help to overcome a major hurdle of data annotation for widespread adoption of deep learning based site specific weed management approaches.

One limitation of our approach is that it relies on the availability of large and diverse datasets for training. While our method is effective in classifying weeds in long-tailed datasets, its performance may be limited by the size and diversity of the available data. In future work, it would be interesting to explore methods for data augmentation or transfer learning to improve the performance of our approach in scenarios where data is limited.

Another limitation of our approach is that it is currently designed for weed classification in long-tailed datasets. While our method shows promising results in this specific application, it remains to be seen how well it generalizes to other classification tasks with different data distributions. In future work, it would be interesting to evaluate the performance of our approach on other classification tasks and explore methods for adapting it to different data distributions.

Another potential direction for future work is to explore the integration of our method with other components of an automated weed management system, such as robotic platforms for weed management that our team has previously developed \citep{calvert2021robotic}. Additionally, further research could investigate the use of our method in other domains such as medical imaging where long-tailed data distributions are common.

\section{Conclusions}  \label{secconc}
This paper introduced Weed Contrastive Learning through visual Representation (WeedCLR), a novel method for weed classification in long-tailed datasets, which leverages self-supervised learning to extract meaningful visual features from images of weeds. The utilization of a class-optimized loss function with Von Neumann Entropy of deep representation significantly improved classification accuracy.

The findings from this study have advanced the current state-of-the-art by demonstrating WeedCLR's ability to discern visual characteristics crucial for weed identification, even in the presence of long-tailed data distributions and without the requirement for laborious human annotation of images. This has substantial implications for the field of automated weed control systems, as it unlocks self-supervised learning for weed recognition, reducing reliance on manual labor and increasing the speed of implementation in new weed and crop scenarios for site specific weed management.

The current findings suggest potential new research directions in improving the accuracy and efficiency of weed classification tasks under varying environmental conditions. They also open up possibilities for integrating WeedCLR into existing automated weed control systems and exploring its effectiveness in different agricultural contexts. This study contributes to the body of knowledge by providing a promising new method for weed classification in long-tailed datasets, thereby paving the way for more effective and efficient weed management systems.


\subsection{CO2 Emission Related to Experiments}
Experiments were conducted using a private infrastructure, which has a carbon efficiency of 0.432 kgCO$_2$eq/kWh. A cumulative of 850 hours of computation was performed on hardware of type RTX 2080 Ti (TDP of 250W).
Total emissions are estimated to be 91.8 kgCO$_2$eq of which 0 percent was directly offset.
Estimations were conducted using the \href{https://mlco2.github.io/impact#compute}{Machine Learning Impact calculator} presented in \citep{Lacoste2019QuantifyingLearning}.

\section*{Acknowledgement}
This research is funded by the partnership between the Australian Government's Reef Trust and the Great Barrier Reef Foundation.

\section*{Additional Information}
\textbf{Competing interests} The authors declare no competing interests.\\


	
\printcredits

\bibliographystyle{cas-model2-names}

\bibliography{references}
	
\end{document}